\documentclass[runningheads]{llncs}

\usepackage{eccv}

\usepackage{eccvabbrv}

\usepackage{graphicx}
\usepackage{booktabs}

\usepackage[accsupp]{axessibility}  

\usepackage[pagebackref,breaklinks,colorlinks,citecolor=eccvblue]{hyperref}

\usepackage{orcidlink}

\usepackage{multirow}
\usepackage{color, colortbl}
\usepackage{subcaption}
\usepackage{tabu}
\usepackage{pifont}
\usepackage{makecell}
\usepackage{paralist}
\usepackage{threeparttable}
\usepackage{enumitem}
\usepackage{hhline}
\usepackage{scalerel,xparse}
\usepackage{wrapfig}

\definecolor{Gray}{gray}{0.5}
\definecolor{LGray}{gray}{0.9}
\definecolor{darkblue}{RGB}{94,110,186}
\definecolor{darkGreen}{RGB}{92, 148, 110}
\definecolor{myblue}{RGB}{14, 121, 178}
\definecolor{myred}{RGB}{192, 0, 0}

\newcommand{\blue}[1]{\textcolor{blue}{#1}}

\newcommand{\gray}[1]{\textcolor{gray}{#1}}

\newcommand{\myred}[1]{\textcolor{myred}{#1}}
\newcommand{\red}[1]{\textcolor{red}{#1}}

\newcommand{\cmark}{\ding{51}}%
\newcommand{\xmark}{\ding{55}}%

\newcommand\blfootnote[1]{%
\begingroup
\renewcommand\thefootnote{}\footnote{#1}%
\addtocounter{footnote}{-1}%
\endgroup
}

\def\ModelName{VideoMamba}

\def\logo{\makebox[22pt][l]{\raisebox{-0.9ex}{\includegraphics[height=30pt]{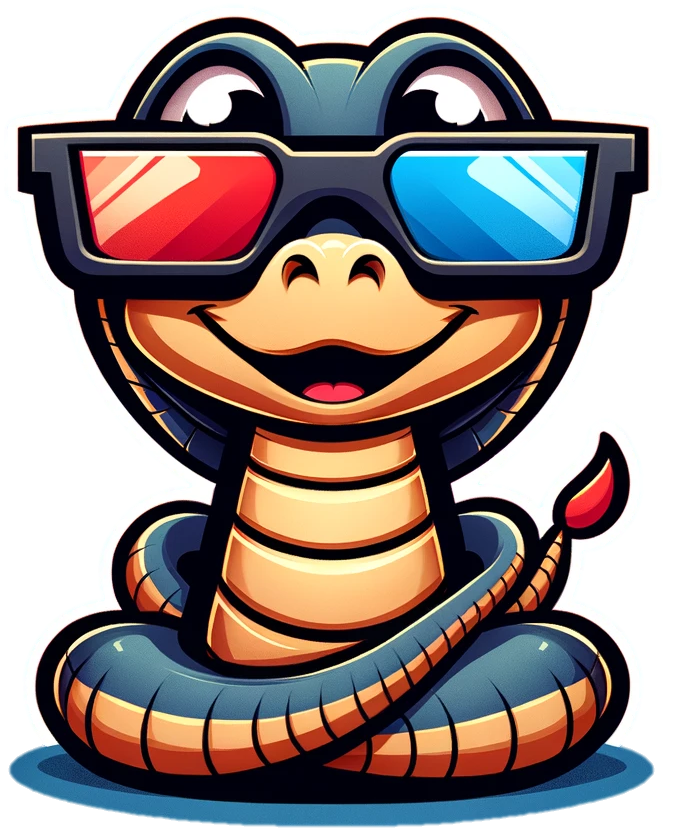}}}\hspace{5pt}}

\begin{document}

\title{\logo\ModelName: State Space Model for Efficient Video Understanding}

\titlerunning{VideoMamba}

\author{Kunchang Li\inst{2,3,1\spadesuit} \and
Xinhao Li\inst{4,1\spadesuit} \and
Yi Wang\inst{1\heartsuit} \and
Yinan He\inst{1} \\
Yali Wang\inst{2,1\heartsuit} \and
Limin Wang\inst{4,1\heartsuit} \and
Yu Qiao\inst{1\heartsuit}}

\authorrunning{K. Li et al.}

\institute{
OpenGVLab, Shanghai AI Laboratory \and
Shenzhen Institute of Advanced Technology, Chinese Academy of Sciences \and
University of Chinese Academy of Sciences \and
State Key Laboratory for Novel Software Technology, Nanjing University\\
\url{https://github.com/OpenGVLab/VideoMamba}
}

\maketitle

\vspace{-0.5cm}
\begin{abstract}

\blfootnote{$\spadesuit$ Interns at Shanghai AI Laboratory. \ \ $\heartsuit$ Corresponding authors.}
Addressing the dual challenges of local redundancy and global dependencies in video understanding,
this work innovatively adapts the Mamba to the video domain.
The proposed \ModelName\  overcomes the limitations of existing 3D convolution neural networks and video transformers.
Its linear-complexity operator enables efficient long-term modeling, 
which is crucial for high-resolution long video understanding. 
Extensive evaluations reveal \ModelName's four core abilities: 
(1) \textit{Scalability} in the visual domain without extensive dataset pretraining, 
thanks to a novel self-distillation technique; 
(2) \textit{Sensitivity} for recognizing short-term actions even with fine-grained motion differences; 
(3) \textit{Superiority} in long-term video understanding, showcasing significant advancements over traditional feature-based models; 
and (4) \textit{Compatibility} with other modalities, 
demonstrating robustness in multi-modal contexts. 
Through these distinct advantages, 
\ModelName\  sets a new benchmark for video understanding, 
offering a scalable and efficient solution for comprehensive video understanding.
All the code and models are available.

\end{abstract}

\section{Introduction}
\label{sec:intro}

The core objective for video understanding lies in mastering spatiotemporal representations, 
which inherently presents two formidable challenges: 
the large spatiotemporal redundancy within short video clips,
and the complex spatiotemporal dependencies among long contexts.
Although the once-dominant 3D convolutional neural networks (CNNs)\cite{c3d,i3d,slowfast} and video transformers\cite{timesformer,vivit}, 
effectively tackle one of the challenges mentioned by leveraging either local convolution or long-range attention, 
they fall short in addressing both simultaneously.
UniFormer~\cite{uniformer} attempts to integrate the advantages of both methods, 
but it struggles with modeling long videos,
which has been the major trend in recent research on video understanding~\cite{gemini,lwm} and generation~\cite{sora,vlogger}.

\begin{figure}[t]
    \centering
    \vspace{-0.3cm}
    \includegraphics[width=1.0\textwidth]{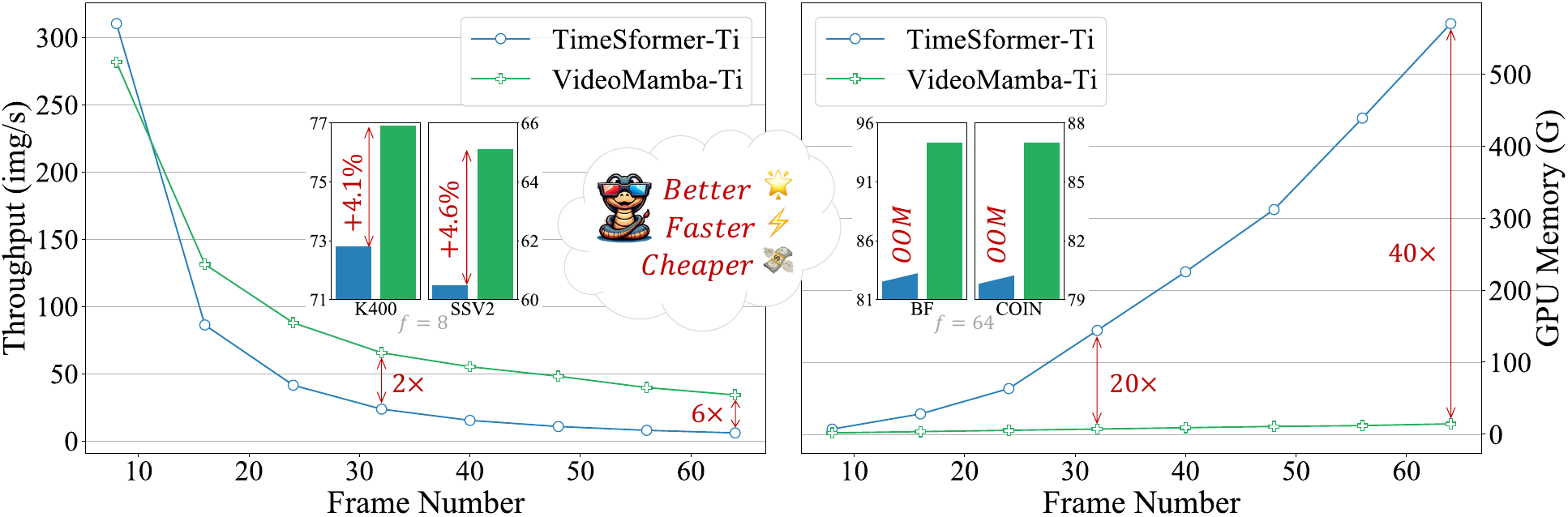}
    \vspace{-0.6cm}
    \caption{\textbf{Comparisons of throughput and memory.}
    The TimeSformer-Ti~\cite{timesformer} is built based on DeiT-Ti~\cite{deit} with joint spatiotemporal attention.
    All the input frames are sized to 224$\times$224.
    The testing is conducted on an NVIDIA A100-80G GPU, utilizing PyTorch 2.1 and CUDA 11.8, with a batch size of 128.
    Our \ModelName\  is \textbf{\textit{better, faster and cheaper}} for both short-term and long-term video understanding.
    } 
    \label{fig:comparison_memory_speed}
    \vspace{-0.3cm}
\end{figure}

The emergence of low-cost operators such as S4~\cite{ssms4}, RWKV~\cite{rwkv}, and RetNet~\cite{retnet} in the NLP domain, 
has carved a novel pathway for the vision model. 
Mamba~\cite{gu2023mamba} stands out with its selective state space model (SSM), 
striking a balance between maintaining linear complexity and facilitating long-term dynamic modeling. 
This innovation has spurred its adoption in vision tasks, 
as evidenced by Vision Mamba~\cite{vim} and VMamba~\cite{vmamba}, 
which leverage multi-directional SSMs for enhanced 2D image processing.
These models rival attention-based architectures in performance while offering a significant reduction in memory usage. 
Given the inherently longer sequences produced by video,
a natural question arises:
\textbf{\textit{Can Mamba work well for video understanding?}}

Inspired by this, 
we introduce \ModelName, 
a purely SSM-based model tailored for video understanding. 
\ModelName\ harmoniously merges the strengths of convolution and attention in vanilla ViT~\cite{vit} style.
It offers a linear-complexity method for dynamic spatiotemporal context modeling, 
ideal for high-resolution long videos. 
The related evaluation focuses on \ModelName's four key abilities:

\textbf{(1) \textit{Scalability in the Visual Domain}:} 
We examine \ModelName's scalability and find that, while the pure Mamba model tends to overfit as it scales, 
our introduction of a simple yet effective self-distillation strategy allows \ModelName\ to achieve remarkable performance enhancements as the model and input sizes increase,
without the need for large-scale dataset pretraining.

\textbf{(2) \textit{Sensitivity for Short-term Action Recognition}:} 
Our analysis extends to assessing \ModelName's capability to accurately distinguish short-term actions,
especially those with fine-grained motion differences, 
\textit{e.g.}, 
opening and closing. 
The findings reveal \ModelName's superior performance over existing attention-based models~\cite{timesformer,vivit,video_swin}.
More importantly,
it is also suitable for masked modeling,
which further enhances its temporal sensitivity. 

\textbf{(3) \textit{Superiority in Long-term Video Understanding}:} 
We then assess \ModelName's prowess in interpreting long videos.
It showcases remarkable superiority over conventional feature-based methods~\cite{vis4mer,distant} through end-to-end training. 
Notably, 
\ModelName\ operates 6$\times$ faster than TimeSformer~\cite{timesformer} and demands 40$\times$ less GPU memory for 64-frame videos (see Fig. \ref{fig:comparison_memory_speed}).

\textbf{(4) \textit{Compatibility with Other Modalities}:} 
Lastly, 
we assess \ModelName's adaptability with other modalities. 
Results in video-text retrievals show its improved performance than ViT, 
particularly in long videos with complex scenarios.
This underscores its robustness and multi-modal integration capacity.

In conclusion, 
our in-depth experiments reveal \ModelName's immense potential in understanding both short-term (K400~\cite{k400} and SthSthV2~\cite{sth}) and long-term (Breakfast~\cite{breakfast}, COIN~\cite{coin}, and LVU~\cite{lvu}) video contents. 
Given its efficiency and effectiveness, 
\ModelName\ is poised to become a cornerstone in the realm of long-video comprehension.
All the code and models are open-sourced to foster future research endeavors.
\section{Related Works}

\subsection{State Space Models}

Recently, the State Space Models (SSMs) have shown significant effectiveness of state space transformation in capturing the dynamics and dependencies of language sequences.
~\cite{ssms4} introduces a structured state-space sequence model (S4), specifically designed to model long-range dependencies, boasting the advantage of linear complexity. 
Based on it,
various models have been developed
(\textit{e.g.}, S5~\cite{ssms5}, H3~\cite{h3layer} and GSS~\cite{mehta2022long}),
and Mamba~\cite{gu2023mamba} distinguishes itself 
by introducing a data-dependent SSM layer and a selection mechanism using parallel scan~(S6).
Compared to transformers~\cite{lu2019vilbert,gpt3} based on quadratic-complexity attention, 
Mamba excels at processing long sequences with linear complexity.

In the vision domain,
\cite{ssms4} first applies SSM in pixel-level image classification,
and \cite{vis4mer} uses S4 to handle the long-range temporal dependencies for movie clip classification.
Besides,
the great potential of Mamba motivates a series of works~\cite{vim,vmamba,liang2024pointmamba,vivim,guo2024mambair,he2024pan,wang2024graph},
which demonstrates Mamba's better performances and higher GPU efficiency than Transformer on visual downstream tasks like object detection and semantic segmentation. 
Different from the previous works,
our \ModelName\ is a purely SSM-based video model,
showcasing great efficiency and effectiveness for both short-term and long-term video understanding.

\subsection{Video Understanding}

Video understanding stands as a cornerstone in the domain of computer vision, 
whose significance is further amplified by the burgeoning growth of short video platforms. 
To bolster this field, 
numerous datasets equipped with extensive data and meticulous human annotations have been developed, aiming to enhance human action recognition capabilities. 
Notable examples include UCF101~\cite{ucf101} and Kinetics dataset~\cite{k400,k600,k700}, 
which have played pivotal roles in benchmarking progress. 
Furthermore, other datasets~\cite{activitynet,thumos,charades_sta,ava,finegym,liu2022fineaction} provide annotated activity videos tailored for action localization,
fostering deeper research into human activities.
Beyond action recognition, 
the advent of large-scale video-text datasets~\cite{msrvtt,msvd,activitynet_qa,youcook2, miech2019howto100m,internvid} extends the utility of video understanding into the realm of multi-modality tasks,
such as video captioning, retrieval and question answering,
thereby broadening the application spectrum. 

As for the architecture,
it has evolved from using CNN which extracts features from video frames, to more advanced techniques.
Initially, 3D CNNs~\cite{c3d,i3d,x3d,r(2+1)d} expanded the traditional 2D CNN architecture to capture videos' spatio-temporal information. 
Two-Stream~\cite{two_stream}, which combines spatial and temporal streams, 
TSN~\cite{tsn}, which proposes sparse sampling,
and SlowFast~\cite{slowfast}, which uses parallel networks to capture semantics and rapid movements, further enhance action recognition capacity. 
The introduction of attention-based models~\cite{timesformer,vivit,motionformer,stam,morphmlp}, like TimeSformer~\cite{timesformer} and ViViT~\cite{vivit}, 
marked a significant advancement by effectively capturing long-range dependencies within video sequences, 
enhancing temporal relationship understanding. Recent developments~\cite{uniformer,uniformerv2,Wang2022InternVideoGV,video_swin} have focused on accurate video transformer, with innovations like the VideoSwin's window attention~\cite{video_swin} and the UniFormer's integration of convolution and self-attention mechanisms~\cite{uniformer}, 
aiming to balance computational efficiency with performance. 
Despite these models' achievements in various tasks, 
they often come with high computational costs for long sequences. 
In contrast,
our \ModelName\  introduces a linear-complexity operator for efficient long-term modeling, outperforming existing methods with faster speed and lower GPU consumption.

\section{Method}
\label{sec:videomamba}

\subsection{Preliminaries}

\noindent\textbf{SSM for 1D sequence.}
State Space Models (SSMs) are conceptualized based on continuous systems that map a 1D function or sequence, $x(t) \in \mathbb{R}^L \rightarrow y(t) \in \mathbb{R}^L $ through a hidden state  $h(t) \in \mathbb{R}^N$. Formally, SSMs employ the following ordinary differential equation (ODE) to model the input data:
\begin{align} 
h'(t) &= {\mathbf A}h(t) + {\mathbf B}x(t), \\
y(t) &= {\mathbf C}h(t), 
\end{align}
where ${\mathbf A} \in \mathbb{R}^{N\times N}$ represents the system's evolution matrix, and ${\mathbf B} \in \mathbb{R}^{N\times 1}, {\mathbf C} \in \mathbb{R}^{ N\times 1}$ are the projection matrices. This continuous ODE is approximated through discretization in modern SSMs. Mamba~\cite{gu2023mamba} is one of the discrete versions of the continuous system, which includes a timescale parameter ${\mathbf \Delta}$ to transform the continuous parameters ${\mathbf A}, {\mathbf B}$ to their discrete counterparts $\overline{{\mathbf A}}, \overline{{\mathbf B}}$. The transformation typically employs the zero-order hold (ZOH) method, defined by:
\begin{align} 
\overline{{\mathbf A}} &= \exp({\mathbf \Delta \mathbf A}), \\
\overline{{\mathbf B}} &= ({\mathbf \Delta \mathbf A})^{-1} (\exp({\mathbf \Delta \mathbf A}) - {\mathbf I}) \cdot {\mathbf \Delta \mathbf B} \\
h_t &= \overline{{\mathbf A}} h_{t-1} + \overline{{\mathbf B}} x_t, \\
y_t &= {\mathbf C}h_t.
\end{align}

\begin{figure}[t]
    \centering
    \vspace{-0.3cm}
    \includegraphics[width=1.0\textwidth]{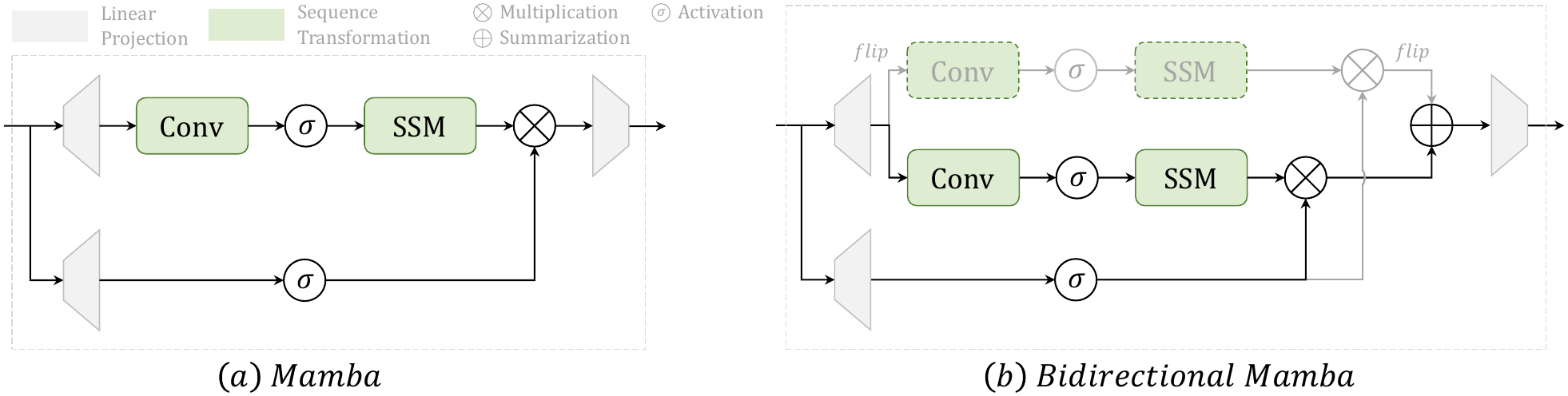}
    \vspace{-0.7cm}
    \caption{
    \textbf{Mamba blocks for 1D~\cite{gu2023mamba} and 2D~\cite{vim} sequence.} 
    We omit the initial normalization and the final residual for simplification.
    }
    \label{fig:mamba_block}
    \vspace{-0.3cm}
\end{figure}

Contrary to traditional models that primarily rely on linear time-invariant SSMs, Mamba distinguishes itself by implementing a Selective Scan Mechanism (S6) as its core SSM operator. Within S6, the parameters ${\mathbf B} \in \mathbb{R}^{B\times L \times N}$, ${\mathbf C} \in \mathbb{R}^{B\times L \times N}$, and ${\mathbf \Delta} \in \mathbb{R}^{B \times L \times D}$ are directly derived from the input data $x \in \mathbb{R}^{B \times L \times D}$, indicating an intrinsic capacity for contextual sensitivity and adaptive weight modulation.
Fig. \ref{fig:mamba_block}\red{a} shows the details of the Mamba block.

\noindent\textbf{Bidirectional SSM for Vision.}
The original Mamba block, 
designed for 1D sequences, 
falls short for visual tasks requiring spatial awareness. 
Building on this, 
Vision Mamba introduces a bidirectional Mamba (B-Mamba) block in Fig. \ref{fig:mamba_block}\red{b},
which adapts bidirectional sequence modeling for vision-specific applications. 
This block processes flattened visual sequences through simultaneous forward and backward SSMs, enhancing its capacity for spatially-aware processing. 
In this work, 
we extend the B-Mamba block for 3D video understanding.

\begin{figure}[t]
    \centering
    \includegraphics[width=1.0\textwidth]{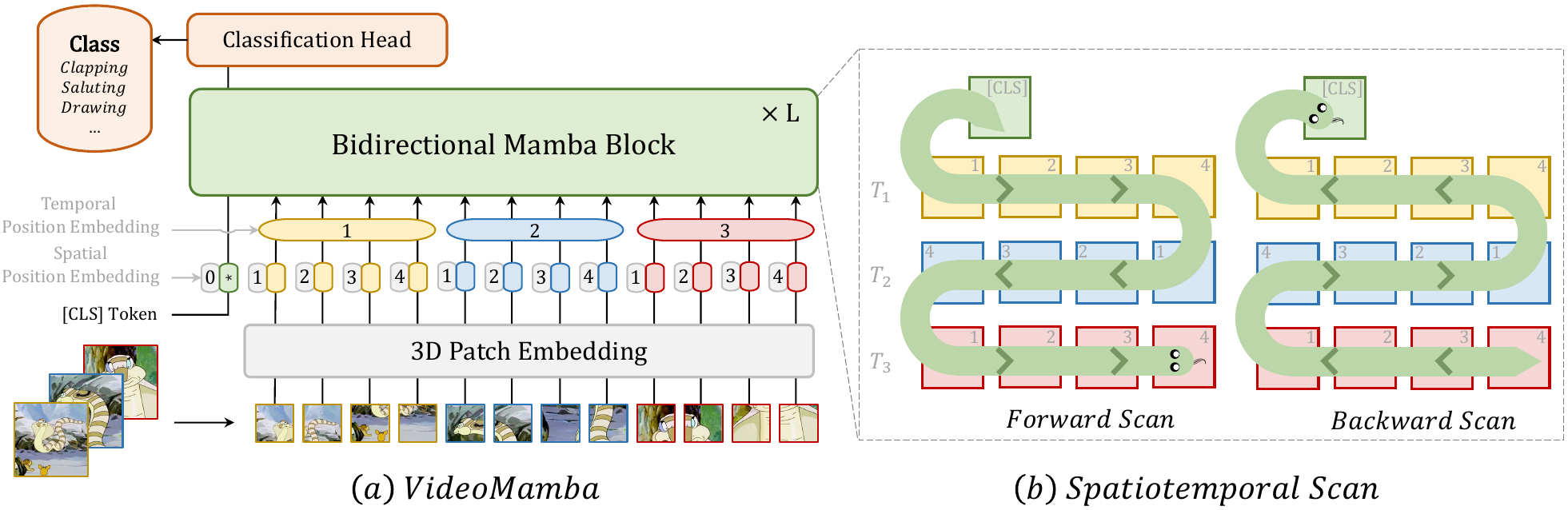}
    \vspace{-0.7cm}
    \caption{
    \textbf{Framework of \ModelName.} 
    We strictly follow the architecture of vanilla ViT~\cite{vit}, and adapt the bidirectional mamba block\cite{vim} for 3D video sequences.
    }
    \label{fig:framework}
    \vspace{-0.3cm}
\end{figure}

\subsection{\ModelName}

\noindent\textbf{Overview.}
Fig.~\ref{fig:framework} illustrates the overall framework of \ModelName.
Specifically,
we first use 3D convolution (\textit{i.e.}, 1$\times$16$\times$16) to project the input videos $\mathbf{X}^{v}\in \mathbb{R}^{3\times T\times H\times W}$ 
into $L$ non-overlapping spatiotemporal patches $\mathbf{X}^{p}\in \mathbb{R}^{L\times C}$,
where $L$$=$$t$$\times$$h$$\times$$w$ ($t$$=$$T$, $h$$=$$\frac{H}{16}$, and $w$$=$$\frac{W}{16}$).
The sequence of tokens input to the following \ModelName\  encoder is
\begin{align}
    \mathbf{X} ={}& \left[ \mathbf{X}_{cls}, \mathbf{X} \right] + \mathbf{p}_{s} + \mathbf{p}_{t},
\end{align}
where $\mathbf{X}_{cls}$ is a learnable classification token that is prepended to the start of the sequence.
Following previous works~\cite{vit,vivit,timesformer},
we added a learnable spatial position embedding $\mathbf{p}_{s} \in \mathbb{R}^{(hw+1) \times C}$ and the extra temporal one $\mathbf{p}_{t} \in \mathbb{R}^{t \times C}$ to retain the spatiotemporal position information,
since the SSM modeling is sensitive to token position.
The tokens $\mathbf{X}$ are then passed through by $L$ stacked B-Mamba blocks,
and the representation of ${\rm [CLS]}$ token at the final layer is processed by normalization and linear layer for classification.

\begin{figure}[t]
    \centering
    \vspace{-0.3cm}
    \includegraphics[width=1.0\textwidth]{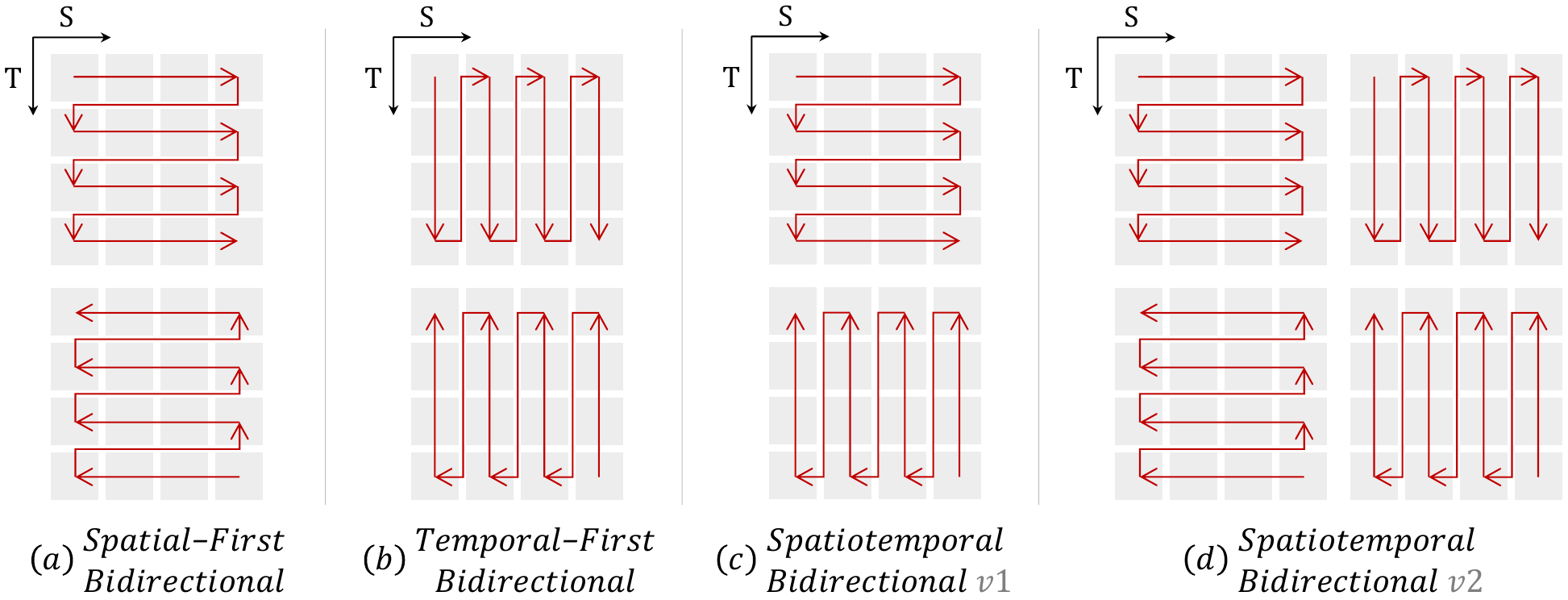}
    \vspace{-0.7cm}
    \caption{
    \textbf{Different scan methods.}
    We omit the [CLS] token for simplification.
    }
    \label{fig:scan}
    \vspace{-0.3cm}
\end{figure}

\vspace{0.05cm}
\noindent\textbf{Spatiotemporal Scan.}
To apply the B-Mamba layer for spatiotemporal input,
we extend the original 2D scan into different bidirectional 3D scans in Fig. \ref{fig:scan}:
(a) \textit{Spatial-First}, organizing spatial tokens by location then stacking them frame by frame;
(b) \textit{Temporal-First}, arranging temporal tokens based on the frame then stacks along the spatial dimension;
(c) \textit{Spatiotemporal}, a hybrid of both \textit{Spatial-First} and \textit{Temporal-First}, with v1 conducting half of them and v2 conducting full of them ($2\times$ computation).
Moreover,
our experiments in Fig. \ref{tab:ablation_scan} demonstrate that the Spatial-First bidirectional scan is the most effective yet simple. 
Thanks to the linear complexity of Mamba,
our \ModelName\  is capable of handling long videos of high resolution efficiently.

\vspace{0.05cm}
\noindent\textbf{Comparison to Vim~\cite{vim} and VMamba~\cite{vmamba}.}
Our \ModelName\  builds upon Vim, 
yet streamlines its architecture by omitting features such as the middle [CLS] token and Rotary Position Embedding (RoPE~\cite{rope}), 
resulting in superior performance on ImageNet-1K with gains of \textbf{+0.8\%} and \textbf{+0.7\%} for Vim-Ti and Vim-S, respectively.
Unlike VMamba, 
which incorporates additional depthwise convolution, 
\ModelName\  strictly follows the ViT design without downsampling layers. 
To counter the overfitting issues observed in VMamba, 
we introduce an effective self-distillation technique outlined in Section~\ref{sec:arch},
demonstrate the isotropic \ModelName's great scalability for image and video tasks.

\vspace{0.05cm}
\noindent\textbf{Comparison to TimeSformer~\cite{timesformer} and ViViT~\cite{vivit}.}
Traditional attention-based models like TimeSformer and ViViT have addressed the self-attention mechanism's quadratic complexity by adopting divided spatiotemporal attention.
Despite being more efficient, 
it introduces additional parameters and underperforms compared to joint attention, 
particularly in scenarios involving masked pretraining~\cite{videomae,umt}. 
In contrast, 
\ModelName\  processes spatiotemporal tokens with linear complexity, 
outperforming TimeSformer on Kinetics-400 by \textbf{+2.6\%} and making significant strides on SthSthV2 with a \textbf{+5.9\%} improvement (see Table \ref{results_k400} and \ref{results_ssv2}). 
Furthermore, 
\ModelName\  achieves a \textbf{6$\times$} increase in processing speed and requires \textbf{40$\times$} less GPU memory for long videos, 
as detailed in Fig. \ref{fig:comparison_memory_speed}, 
demonstrating its efficiency and effectiveness in handling long-video tasks.

\subsection{Architecture}
\label{sec:arch}

\begin{wraptable}{r}{5cm}
  \vspace{-0.8cm}
  \centering
  \setlength\tabcolsep{4pt}
  \resizebox{0.43\textwidth}{!}{
  \begin{tabular}{lrrr}
    \Xhline{1.0pt}
    \small{\textbf{Model}} & \scriptsize{\textbf{\#Depth}} & \scriptsize{\textbf{\#Dim}} & \scriptsize{\textbf{\#Param.}} \\
    \Xhline{0.8pt}
    Tiny & 24 & 192 & 7M \\
    Small & 24 & 384 & 26M \\
    Middle & 32 & 576 & 74M \\
    \gray{Base} & \gray{24} & \gray{768} & \gray{98M}  \\
    \Xhline{1.0pt}
  \end{tabular}
  }
  \vspace{-0.2cm}
  \caption{\textbf{Different model sizes.} \gray{Base model is finally excluded} due to its suboptimization.} 
  \label{tab:model_size}
  \vspace{-0.6cm}
\end{wraptable}

For SSM in the B-Mamba layer,
we adopt the default hyperparameters as in Mamba~\cite{gu2023mamba}.
setting the state dimension and expansion ratio to 16 and 2, respectively.
Following ViT~\cite{vit},
we adjust the depth and embedding dimensions to create models of comparable sizes in Table \ref{tab:model_size}, 
including \ModelName-Ti, \ModelName-S and \ModelName-M. 
However, 
we observe that larger \ModelName\ tends to overfit during our experiments, 
leading to suboptimal performance as illustrated in Fig. \ref{tab:ablation_sd}\red{a}. 
This overfitting issue is not unique to our models but is also found in VMamba~\cite{vmamba}, 
where the optimal performance of VMamba-B was achieved at three-quarters of the total training epochs. 
To counteract the overfitting in larger Mamba models, 
we introduce an effective Self-Distillation strategy,
which uses a smaller and well-trained model as the ``teacher'' to guide the training of the larger ``student'' model. 
The results, depicted in Fig. \ref{tab:ablation_sd}\red{a}, show that this strategy leads to expected better convergence.

\subsection{Masked Modeling}

\begin{figure}[t]
    \centering
    \vspace{-0.3cm}
    \includegraphics[width=1.0\textwidth]{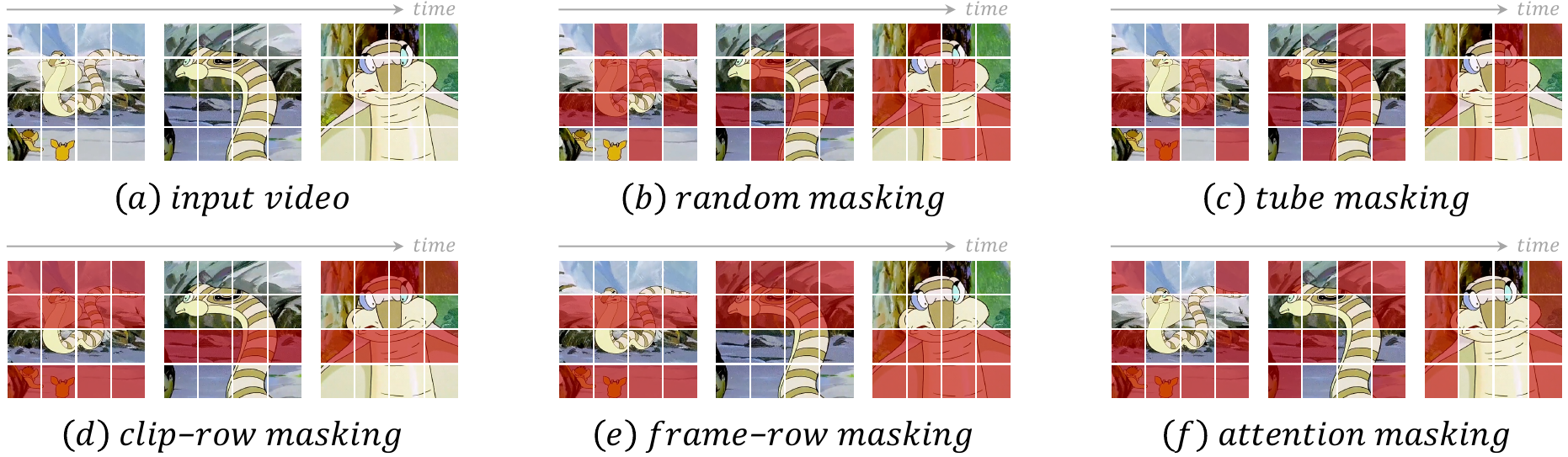}
    \vspace{-0.6cm}
    \caption{
    \textbf{Different masking strategies.}
    Row masking, 
    tailored for \ModelName\ in light of the 1D convolution preceding SSM, 
    enhances performance with continuous tokens. 
    The difference between clip-row and frame-row masking is that the former masks the entire video clip, while the latter masks each frame individually.
    }
    \label{fig:mask}
    \vspace{-0.3cm}
\end{figure}

Recently, 
VideoMAE and ST-MAE~\cite{videomae,st_mae} have showcased the significant benefits of masked modeling in enhancing a model's capability for FINE-GRAINED temporal understanding. 
UMT~\cite{umt} takes this further by introducing an efficient masked alignment technique that yields robust results across single and multi-modal video tasks. 
To augment \ModelName's temporal sensitivity and verify its adaptability with text modalities, 
we adopt a masked alignment approach inspired by UMT. 
Firstly,
\ModelName\  is trained from scratch on video data alone, 
aligning unmasked tokens with those from CLIP-ViT. Subsequently, 
it is integrated with a text encoder and a cross-modal decoder (\textit{i.e.}, BERT~\cite{devlin2018bert}), 
for pretraining on both image-text and video-text datasets.

It's important to note the distinction from UMT, 
which employs multi-layer alignment between the student and teacher models. 
In contrast, 
due to \ModelName's unique architecture (SSM \textit{vs.} Transformer), 
we align only the final outputs. 
Regarding our masking strategy, 
we propose different row masking techniques, 
depicted in Fig.~\ref{fig:mask}, 
tailored to the B-Mamba block's preference for continuous tokens. 
Additionally, 
we explore attention masking to preserve meaningful adjacency among tokens, 
leveraging the inherent strengths of the 1D convolution within the B-Mamba block for improved performance.
\section{Experiments}
\label{sec:exp}

\subsection{Scaling Up}
\begin{figure}[t]
\centering
\vspace{-0.3cm}
\centering
\includegraphics[width=\linewidth]{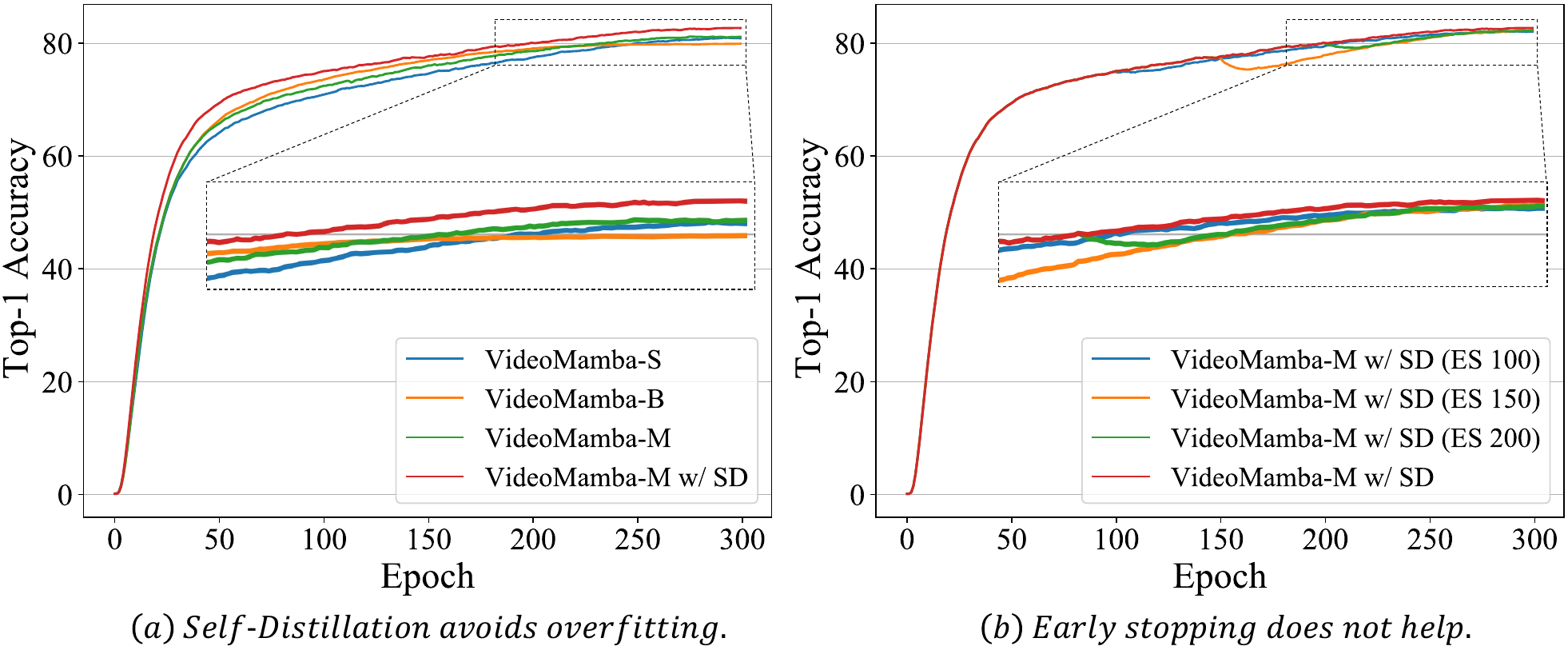} 
\vspace{-0.6cm}
\caption{\textbf{Ablation studies of Self-Distillation and Early Stopping.}
} 
\label{tab:ablation_sd}
\vspace{-0.3cm}
\end{figure}

\noindent\textbf{Dataset and Settings.} 
We first conduct experiments on ImageNet-1K~\cite{imagenet},
which includes 1.28M training images and 50K validation images across 1,000 categories.
For fair comparisons,
we follow most of the training strategies proposed in DeiT~\cite{deit},
but adopt weaker data augmentation for the tiny model variant.
Furthermore,
we adjust the stochastic depth ratio to 0/0.15/0.5 for \ModelName-Ti/S/M.
Our models are trained using the AdamW optimizer paired with a cosine learning rate schedule over 300 epochs. 
The initial 5 epochs serve as a period for linear warm-up. 
Default settings for the learning rate, weight decay, and batch size are 1e-3, 0.05, and 1024, respectively. 
Moreover, 
we use BFloat16 precision during training to enhance stability without relying on EMA.
For the \ModelName-M model, 
we employ a pretrained \ModelName-S model as a ``teacher'' to guide the training process by aligning the final feature maps through L2 loss.
For large resolution ($>$224) fine-tuning,
we use a reduced learning rate (5e-6) and minimal weight decay (1e-8) for 30 epochs.

\vspace{0.05cm}
\noindent\textbf{Effect of Self-Distillation.} 
Fig. \ref{tab:ablation_sd}\red{a} reveals that when trained from scratch,
\ModelName-B tends to overfit more easily and underperforms compared to \ModelName-S, 
whereas \ModelName-M achieves similar performances. 
Fortunately, 
our self-distillation has shown to be effective in achieving the desired optimization with marginal additional computational cost. 
To mitigate teacher's potential overdirection, 
we experimented with early stopping~\cite{early_stopping} in Fig. \ref{tab:ablation_sd}\red{b}, 
although it did not yield beneficial outcomes. 
These findings indicate that self-distillation offers a viable strategy for enhancing the scalability of the Mamba architecture without significant computational overhead.

\begin{table}[tp]
    \vspace{-0.3cm}
    \centering
    \setlength\tabcolsep{6pt}
    \resizebox{0.8\linewidth}{!}{
        \begin{tabular}{l|l|c|r|r|r|c}
        \Xhline{1.0pt}
            \multirow{2}*{\textbf{Arch.}} & \multirow{2}*{\textbf{Model}} & \multirow{2}*{\textit{\textbf{iso.}}} & \textbf{Input} & \textbf{\#Param} & \textbf{FLOPs} & \textbf{IN-1K} \\
            ~ & ~ & ~ & \textbf{Size} & \textbf{(M)} & \textbf{(G)} & \textbf{Top-1}  \\
            \Xhline{0.8pt}
            \multirow{4}{*}{\textbf{\textit{CNN}}} & ConvNeXt-T~\cite{convnext} & \xmark & 224$^2$ & 29 & 4.5 & 82.1 \\
            ~ & ConvNeXt-S~\cite{convnext} & \xmark & 224$^2$ & 50 & 8.7 & 83.1 \\
            ~ & ConvNeXt-B~\cite{convnext} & \xmark & 224$^2$ & 89 & 15.4 & \textbf{83.8} \\
            \hline
            \multirow{4}{*}{\textit{\textbf{Trans.}}}  & SwinT-T~\cite{swin} & \xmark & 224$^2$ & 28 & 4.5 & 81.3 \\
            ~ & Swin-S~\cite{swin} & \xmark & 224$^2$ & 50 & 8.7 & 83.0 \\
            ~ & Swin-B~\cite{swin} & \xmark & 224$^2$ & 88 & 15.4 & 83.5 \\
            \hline
            \multirow{3}{*}{\textit{\textbf{\makecell[l]{CNN+\\SSM}}}}  & VMamba-T~\cite{vmamba} & \xmark & 224$^2$ & 22 & 5.6 & 82.2 \\
            ~ & VMamba-S~\cite{vmamba} & \xmark & 224$^2$ & 44 & 11.2 & 83.5 \\
            ~ & VMamba-B~\cite{vmamba} & \xmark & 224$^2$ & 75 & 18.0 & \underline{83.7} \\
            \Xhline{1.0pt}
            \multirow{2}{*}{\textit{\textbf{CNN}}} & ConvNeXt-S~\cite{convnext} & \cmark & 224$^2$ & 22 & 4.3 & 79.7 \\
            ~ & ConvNeXt-B~\cite{convnext} & \cmark & 224$^2$ & 87 & 16.9 & 82.0 \\
            \hline
            \multirow{4}{*}{\textit{\textbf{Trans.}}}  & DeiT-Ti~\cite{deit} & \cmark & 224$^2$ & 6 & 1.3 & 72.2 \\
            ~ & DeiT-S~\cite{deit} & \cmark & 224$^2$ & 22 & 4.6 & 79.8 \\
            ~ & DeiT-B~\cite{deit} & \cmark & 224$^2$ & 87 & 17.6 & 81.8 \\
            ~ & DeiT-B~\cite{deit} & \cmark & 384$^2$ & 87 & 55.5 & \underline{83.1} \\
            \hline
            \multirow{12}{*}{\textit{\textbf{SSM}}}  & S4ND-ViT-B~\cite{s4nd} & \cmark & 224$^2$ & 89 & - & 80.4 \\
            ~ & Vim-Ti~\cite{vim} & \cmark & 224$^2$ & 7 & 1.1 & 76.1 \\
            ~ & Vim-S~\cite{vim} & \cmark & 224$^2$ & 26 & 4.3 & 80.5 \\
            ~ & \cellcolor{blue!10}{\ModelName-Ti} & \cellcolor{blue!10}{\cmark} & \cellcolor{blue!10}{224$^2$} & \cellcolor{blue!10}{7} & \cellcolor{blue!10}{1.1} & \cellcolor{blue!10}{76.9} \\
            ~ & \cellcolor{blue!10}{\ModelName-Ti} & 
            \cellcolor{blue!10}{\cmark} & \cellcolor{blue!10}{448$^2$} & \cellcolor{blue!10}{7} & \cellcolor{blue!10}{4.3} & \cellcolor{blue!10}{79.3} \\
            ~ & \cellcolor{blue!10}{\ModelName-Ti} & 
            \cellcolor{blue!10}{\cmark} & \cellcolor{blue!10}{576$^2$} & \cellcolor{blue!10}{7} & \cellcolor{blue!10}{7.1} & \cellcolor{blue!10}{79.6} \\
            ~ & \cellcolor{blue!10}{\ModelName-S} & 
            \cellcolor{blue!10}{\cmark} & \cellcolor{blue!10}{224$^2$} & \cellcolor{blue!10}{26} & \cellcolor{blue!10}{4.3} & \cellcolor{blue!10}{81.2} \\
             ~ & \cellcolor{blue!10}{\ModelName-S} & 
            \cellcolor{blue!10}{\cmark} & \cellcolor{blue!10}{448$^2$} & \cellcolor{blue!10}{26} & \cellcolor{blue!10}{16.9} & \cellcolor{blue!10}{83.2} \\
            ~ & \cellcolor{blue!10}{\ModelName-S} & 
            \cellcolor{blue!10}{\cmark} & \cellcolor{blue!10}{576$^2$} & \cellcolor{blue!10}{26} & \cellcolor{blue!10}{28.0} & \cellcolor{blue!10}{83.5} \\
            ~ & \cellcolor{blue!10}{\ModelName-M} & 
            \cellcolor{blue!10}{\cmark} & \cellcolor{blue!10}{224$^2$} & \cellcolor{blue!10}{74} & \cellcolor{blue!10}{12.7} & \cellcolor{blue!10}{82.8} \\
             ~ & \cellcolor{blue!10}{\ModelName-M} & 
            \cellcolor{blue!10}{\cmark} & \cellcolor{blue!10}{448$^2$} & \cellcolor{blue!10}{75} & \cellcolor{blue!10}{50.4} & \cellcolor{blue!10}{83.8} \\
            ~ & \cellcolor{blue!10}{\ModelName-M} & 
            \cellcolor{blue!10}{\cmark} & \cellcolor{blue!10}{576$^2$} & \cellcolor{blue!10}{75} & \cellcolor{blue!10}{83.1} & \cellcolor{blue!10}{\textbf{84.0}} \\
        \Xhline{1.0pt}	
        \end{tabular}
    }
    \caption{\textbf{Comparison with the state-of-the-art on ImageNet.} 
    ``\textit{iso.}'' means isotropic architecture without downsampling layers.
    }
    \label{results_imagenet}
    \vspace{-0.5cm}
\end{table}

\vspace{0.05cm}
\noindent\textbf{Results.}
Table~\ref{results_imagenet} showcases the results on the ImageNet-1K dataset. 
Notably, 
\ModelName-M outperforms other isotropic architectures by significant margins, achieving a \textbf{+0.8\%} improvement over ConvNeXt-B~\cite{convnext} and a \textbf{+2.0\%} increase compared to DeiT-B~\cite{deit}, 
while utilizing fewer parameters. 
Additionally, \ModelName-M holds its ground against non-isotropic backbones that leverage hierarchical features for enhanced performance. 
Given Mamba's efficiency in processing long sequences, 
we further enhance performance by increasing the resolution, achieving a top-1 accuracy of \textbf{84.0\%} with only 74M parameters. 
This remarkable improvement extends to video tasks, as detailed in Section \ref{sec:short_term}, underscoring \ModelName's effectiveness and scalability.

\begin{table}[t]
    \vspace{-0.3cm}
    \centering
    \setlength\tabcolsep{2pt}
    \resizebox{0.95\linewidth}{!}{
        \begin{tabular}{l|l|c|l|r|r|r|cc}
        \Xhline{1.0pt}
            \multirow{2}*{\textbf{Arch.}} & \multirow{2}*{\textbf{Model}} & \multirow{2}*{\textit{\textbf{iso.}}} & \textbf{Extra} & \textbf{Input} & \textbf{\#Param} & \textbf{FLOPs} & \multicolumn{2}{c}{\textbf{K400}} \\
            ~ & ~ & ~ & \textbf{Data} & \textbf{Size} & \textbf{(M)} & \textbf{(G)} & \textbf{Top-1} & \textbf{Top-5} \\
            \Xhline{0.8pt}
             \multicolumn{9}{l}{\textit{\textbf{\blue{Supervised:}} \gray{Those models with extra data are under supervised training.}}} \\
             \multirow{3}{*}{\textit{\textbf{CNN}}} & SlowFast$_{R101+NL}$~\cite{slowfast} & \xmark &  & 80$\times$224$^2$ & 60 & 234$\times$3$\times$10 & 79.8 & 93.9 \\
             ~ & X3D-M~\cite{x3d} & \xmark &  & 16$\times$224$^2$ & 4 & 6$\times$3$\times$10 & 76.0 & 92.3 \\
             ~ & X3D-XL~\cite{x3d} & \xmark &  & 16$\times$312$^2$ & 20 & 194$\times$3$\times$10 & 80.4 & 94.6 \\
             \hline
             \multirow{3}{*}{\textit{\textbf{Trans.}}} & Swin-T~\cite{video_swin} & \xmark & IN-1K & 32$\times$224$^2$ & 28 & 88$\times$3$\times$4 & 78.8 & 93.6 \\
             ~ & Swin-B~\cite{video_swin} & \xmark & IN-1K & 32$\times$224$^2$ & 88 & 88$\times$3$\times$4 & 80.6 & 94.5 \\
             ~ & Swin-B~\cite{video_swin} & \xmark & IN-21K & 32$\times$224$^2$ & 88 & 282$\times$3$\times$4 & \underline{82.7} & \textbf{95.5} \\
             \hline
             \multirow{5}{*}{\textbf{\textit{\makecell[l]{CNN+\\Trans.}}}} & MViTv1-B~\cite{mvit} & \xmark &  & 32$\times$224$^2$ & 37 & 70$\times$1$\times$5 & 80.2 & 94.4 \\
             ~ & MViTv2-S~\cite{mvitv2} & \xmark &  & 16$\times$224$^2$ & 35 & 64$\times$1$\times$5 & 81.0 & 94.6 \\
             ~ & UniFormer-S~\cite{uniformer} & \xmark & IN-1K & 16$\times$224$^2$ & 21 & 42$\times$1$\times$4 & 80.8 & 94.7 \\
             ~ & UniFormer-B~\cite{uniformer} & \xmark & IN-1K & 16$\times$224$^2$ & 50 & 97$\times$1$\times$4 & 82.0 & 95.1 \\
             ~ & UniFormer-B~\cite{uniformer} & \xmark & IN-1K & 32$\times$224$^2$ & 50 & 259$\times$3$\times$4 & \textbf{83.0} & \underline{95.4} \\
             \Xhline{0.8pt}
             \multirow{4}{*}{\textit{\textbf{Trans.}}} & STAM~\cite{stam} & \cmark & IN-21K & 64$\times$224$^2$ & 121 & 1040$\times$1$\times$1 & 79.2 & - \\
             ~ & TimeSformer-L~\cite{timesformer} & \cmark & IN-21K & 96$\times$224$^2$ & 121 & 2380$\times$3$\times$1 & 80.7 & 94.7 \\
             ~ & ViViT-L~\cite{vivit} & \cmark & IN-21K & 16$\times$224$^2$ & 311 & 3992$\times$3$\times$4 & \underline{81.3} & 94.7 \\
             ~ & Mformer-HR~\cite{motionformer} & \cmark & IN-21K & 16$\times$336$^2$ & 311 & 959$\times$3$\times$10 & 81.1 & \underline{95.2} \\
             \hline
             \multirow{9}{*}{\textit{\textbf{SSM}}} & \cellcolor{blue!10}{\ModelName-Ti} & \cellcolor{blue!10}{\cmark} & \cellcolor{blue!10}{IN-1K} & \cellcolor{blue!10}{16$\times$224$^2$} & \cellcolor{blue!10}{7} & \cellcolor{blue!10}{17$\times$3$\times$4} & \cellcolor{blue!10}{78.1} & \cellcolor{blue!10}{93.5} \\
             ~ & \cellcolor{blue!10}{\ModelName-Ti} & \cellcolor{blue!10}{\cmark} & \cellcolor{blue!10}{IN-1K} & \cellcolor{blue!10}{32$\times$224$^2$} & \cellcolor{blue!10}{7} & \cellcolor{blue!10}{34$\times$3$\times$4} & \cellcolor{blue!10}{78.8} & \cellcolor{blue!10}{93.9} \\
             ~ & \cellcolor{blue!10}{\ModelName-Ti} & \cellcolor{blue!10}{\cmark} & \cellcolor{blue!10}{IN-1K} & \cellcolor{blue!10}{64$\times$384$^2$} & \cellcolor{blue!10}{7} & \cellcolor{blue!10}{202$\times$3$\times$4} & \cellcolor{blue!10}{80.3} & \cellcolor{blue!10}{94.8} \\

             ~ & \cellcolor{blue!10}{\ModelName-S} & \cellcolor{blue!10}{\cmark} & \cellcolor{blue!10}{IN-1K} & \cellcolor{blue!10}{16$\times$224$^2$} & \cellcolor{blue!10}{26} & \cellcolor{blue!10}{68$\times$3$\times$4} & \cellcolor{blue!10}{80.8} & \cellcolor{blue!10}{94.8} \\
             ~ & \cellcolor{blue!10}{\ModelName-S} & \cellcolor{blue!10}{\cmark} & \cellcolor{blue!10}{IN-1K} & \cellcolor{blue!10}{32$\times$224$^2$} & \cellcolor{blue!10}{26} & \cellcolor{blue!10}{135$\times$3$\times$4} & \cellcolor{blue!10}{81.5} & \cellcolor{blue!10}{95.2} \\
             ~ & \cellcolor{blue!10}{\ModelName-S} & \cellcolor{blue!10}{\cmark} & \cellcolor{blue!10}{IN-1K} & \cellcolor{blue!10}{64$\times$384$^2$} & \cellcolor{blue!10}{26} & \cellcolor{blue!10}{395$\times$3$\times$4} & \cellcolor{blue!10}{82.7} & \cellcolor{blue!10}{95.6} \\

             ~ & \cellcolor{blue!10}{\ModelName-M} & \cellcolor{blue!10}{\cmark} & \cellcolor{blue!10}{IN-1K} & \cellcolor{blue!10}{16$\times$224$^2$} & \cellcolor{blue!10}{74} & \cellcolor{blue!10}{202$\times$3$\times$4} & \cellcolor{blue!10}{81.9} & \cellcolor{blue!10}{95.4} \\
             ~ & \cellcolor{blue!10}{\ModelName-M} & \cellcolor{blue!10}{\cmark} & \cellcolor{blue!10}{IN-1K} & \cellcolor{blue!10}{32$\times$224$^2$} & \cellcolor{blue!10}{74} & \cellcolor{blue!10}{403$\times$3$\times$4} & \cellcolor{blue!10}{82.4} & \cellcolor{blue!10}{95.7} \\
             ~ & \cellcolor{blue!10}{\ModelName-M} & \cellcolor{blue!10}{\cmark} & \cellcolor{blue!10}{IN-1K} & \cellcolor{blue!10}{64$\times$384$^2$} & \cellcolor{blue!10}{74} & \cellcolor{blue!10}{2368$\times$3$\times$4} & \cellcolor{blue!10}{\textbf{83.3}} & \cellcolor{blue!10}{\textbf{96.1}} \\
             
             \Xhline{1.0pt}
             \multicolumn{9}{l}{\textit{\textbf{\blue{Self-supervised:}} \gray{For UMT, the CLIP-400M is used in pretrained teacher.}}} \\
             \multirow{5}{*}{\textit{\textbf{Trans.}}} & BEVT-B$_{800e}$~\cite{bevt} & \xmark & IN-1K & 32$\times$224$^2$ & 88 & 282$\times$3$\times$4 & 81.1 & - \\
             \cline{2-9}
             ~ & ST-MAE-B$_{1600e}$~\cite{st_mae} & \cmark &  & 16$\times$224$^2$ & 87 & 180$\times$3$\times$7 & 81.3 & 94.9 \\
             ~ & VideoMAE-S$_{2400e}$~\cite{videomae} & \cmark &  & 16$\times$224$^2$ & 22 & 57$\times$3$\times$5 & 79.0 & 93.8 \\
             ~ & VideoMAE-B$_{1600e}$~\cite{videomae} & \cmark &  & 16$\times$224$^2$ & 87 & 180$\times$3$\times$5 & 81.5 & 95.1 \\
             ~ & UMT-B$_{800e}$~\cite{umt} & \cmark & \gray{CLIP-400M} & 8$\times$224$^2$ & 87 & 180$\times$3$\times$5 & \textbf{85.7} & \textbf{97.0} \\

             \hline
             \multirow{4}{*}{\textit{\textbf{SSM}}} & \cellcolor{blue!10}{\ModelName-M$_{800e}$} & \cellcolor{blue!10}{\cmark} & \cellcolor{blue!10}{\gray{CLIP-400M}} & \cellcolor{blue!10}{8$\times$224$^2$} & \cellcolor{blue!10}{74} & \cellcolor{blue!10}{101$\times$3$\times$4} & \cellcolor{blue!10}{82.0} & \cellcolor{blue!10}{95.4} \\
             ~ & \cellcolor{blue!10}{\ModelName-M$_{800e}$} & \cellcolor{blue!10}{\cmark} & \cellcolor{blue!10}{\gray{CLIP-400M}} & \cellcolor{blue!10}{16$\times$224$^2$} & \cellcolor{blue!10}{74} & \cellcolor{blue!10}{202$\times$3$\times$4} & \cellcolor{blue!10}{83.4} & \cellcolor{blue!10}{95.9} \\
             ~ & \cellcolor{blue!10}{\ModelName-M$_{800e}$} & \cellcolor{blue!10}{\cmark} & \cellcolor{blue!10}{\gray{CLIP-400M}} & \cellcolor{blue!10}{32$\times$224$^2$} & \cellcolor{blue!10}{74} & \cellcolor{blue!10}{403$\times$3$\times$4} &\cellcolor{blue!10}{83.9} & \cellcolor{blue!10}{96.2} \\
             ~ & \cellcolor{blue!10}{\ModelName-M$_{800e}$} & \cellcolor{blue!10}{\cmark} & \cellcolor{blue!10}{\gray{CLIP-400M}} & \cellcolor{blue!10}{64$\times$384$^2$} & \cellcolor{blue!10}{74} & \cellcolor{blue!10}{2368$\times$3$\times$4} &\cellcolor{blue!10}{\underline{85.0}} & \cellcolor{blue!10}{\underline{96.9}} \\
        \Xhline{1.0pt}	
        \end{tabular}
    }
    \caption{\textbf{Comparison with the state-of-the-art on scene-related Kinetics-400.} 
    ``\textit{iso.}'' means isotropic architecture without downsampling layers.
    Masked modeling~\cite{umt} also works for Mamba,
    but the inconsistent architecture leads to inferior alignment.
    }
    \label{results_k400}
    \vspace{-0.6cm}
\end{table}  

\begin{table}[t]
    \vspace{-0.3cm}
    \centering
    \setlength\tabcolsep{2pt}
    \resizebox{0.96\linewidth}{!}{
        \begin{tabular}{l|l|c|l|r|r|r|cc}
        \Xhline{1.0pt}
            \multirow{2}*{\textbf{Arch.}} & \multirow{2}*{\textbf{Model}} & \multirow{2}*{\textit{\textbf{iso.}}} & \textbf{Extra} & \textbf{Input} & \textbf{\#Param} & \textbf{FLOPs} & \multicolumn{2}{c}{\textbf{SSV2}} \\
            ~ & ~ & ~ & \textbf{Data} & \textbf{Size} & \textbf{(M)} & \textbf{(G)} & \textbf{Top-1} & \textbf{Top-5} \\
            \Xhline{0.8pt}
             \multicolumn{9}{l}{\textit{\textbf{\blue{Supervised:}} \gray{Those models with extra data are under supervised training.}}} \\
             \multirow{3}{*}{\textit{\textbf{CNN}}} & SlowFast$_{R101}$~\cite{slowfast} & \xmark & K400 & 32$\times$224$^2$ & 53 & 106$\times$3$\times$1 & 63.1 & 87.6 \\
             ~ & CT-Net$_{R50}$~\cite{ct_net} & \xmark & IN-1K & 16$\times$224$^2$ & 21 & 75$\times$1$\times$1 & 64.5 & 89.3 \\
             ~ & TDN$_{R50}$~\cite{tdn} & \xmark & IN-1K & 16$\times$224$^2$ & 26 & 75$\times$1$\times$1 & 65.3 & 91.6 \\
             \hline
             \multirow{1}{*}{\textit{\textbf{Trans.}}} & Swin-B~\cite{video_swin} & \xmark & K400 & 32$\times$224$^2$ & 89 & 88$\times$3$\times$1 & 69.6 & 92.7 \\
             \hline
             \multirow{6}{*}{\textbf{\textit{\makecell[l]{CNN+\\Trans.}}}} & MViTv1-B~\cite{mvit} & \xmark & K400 & 16$\times$224$^2$ & 37 & 71$\times$3$\times$1 & 64.7 & 89.2 \\
             ~ & MViTv1-B~\cite{mvit} & \xmark & K400 & 32$\times$224$^2$ & 37 & 170$\times$3$\times$1 & 67.1 & 90.8 \\
             ~ & MViTv2-S~\cite{mvitv2} & \xmark & K400 & 16$\times$224$^2$ & 35 & 65$\times$3$\times$1 & 68.2 & 91.4 \\
             ~ & MViTv2-B~\cite{mvitv2} & \xmark & K400 & 32$\times$224$^2$ & 51 & 225$\times$3$\times$1 & \textbf{70.5} & \underline{92.7} \\
             ~ & UniFormer-S~\cite{uniformer} & \xmark & \scriptsize{IN-1K+K400} & 16$\times$224$^2$ & 21 & 42$\times$3$\times$1 & 67.7 & 91.4 \\
             ~ & UniFormer-B~\cite{uniformer} & \xmark & \scriptsize{IN-1K+K400} & 16$\times$224$^2$ & 50 & 97$\times$3$\times$1 & \underline{70.4} & \textbf{92.8} \\
             \Xhline{0.8pt}
             \multirow{3}{*}{\textit{\textbf{Trans.}}} & TimeSformer-HR~\cite{timesformer} & \cmark & IN-21K & 16$\times$224$^2$ & 121 & 1703$\times$3$\times$1 & 62.5 & - \\
             ~ & ViViT-L~\cite{vivit} & \cmark & \scriptsize{IN-21K+K400} & 16$\times$224$^2$ & 311 & 3992$\times$3$\times$4 & 65.4 & 89.8 \\
             ~ & Mformer-HR~\cite{motionformer} & \cmark & \scriptsize{IN-21K+K400} & 16$\times$336$^2$ & 311 & 1185$\times$3$\times$1 & \underline{68.1} & \underline{91.2} \\
             \hline
             \multirow{9}{*}{\textit{\textbf{SSM}}} & \cellcolor{blue!10}{\ModelName-Ti} & \cellcolor{blue!10}{\cmark} & \cellcolor{blue!10}{IN-1K} & \cellcolor{blue!10}{8$\times$224$^2$} & \cellcolor{blue!10}{7} & \cellcolor{blue!10}{9$\times$3$\times$2} & \cellcolor{blue!10}{65.1} & \cellcolor{blue!10}{89.1} \\
             ~ & \cellcolor{blue!10}{\ModelName-Ti} & \cellcolor{blue!10}{\cmark} & \cellcolor{blue!10}{IN-1K} & \cellcolor{blue!10}{16$\times$224$^2$} & \cellcolor{blue!10}{7} & \cellcolor{blue!10}{17$\times$3$\times$2} & \cellcolor{blue!10}{66.0} & \cellcolor{blue!10}{89.6} \\
             ~ & \cellcolor{blue!10}{\ModelName-Ti} & \cellcolor{blue!10}{\cmark} & \cellcolor{blue!10}{IN-1K} & \cellcolor{blue!10}{16$\times$288$^2$} & \cellcolor{blue!10}{7} & \cellcolor{blue!10}{28$\times$3$\times$2} & \cellcolor{blue!10}{66.2} & \cellcolor{blue!10}{90.0} \\

             ~ & \cellcolor{blue!10}{\ModelName-S} & \cellcolor{blue!10}{\cmark} & \cellcolor{blue!10}{IN-1K} & \cellcolor{blue!10}{8$\times$224$^2$} & \cellcolor{blue!10}{26} & \cellcolor{blue!10}{34$\times$3$\times$2} & \cellcolor{blue!10}{66.6} & \cellcolor{blue!10}{90.4} \\
             ~ & \cellcolor{blue!10}{\ModelName-S} &\cellcolor{blue!10}{\cmark} & \cellcolor{blue!10}{IN-1K} & \cellcolor{blue!10}{16$\times$224$^2$} & \cellcolor{blue!10}{26} & \cellcolor{blue!10}{68$\times$3$\times$2} & \cellcolor{blue!10}{67.6} & \cellcolor{blue!10}{90.9} \\
             ~ & \cellcolor{blue!10}{\ModelName-S} &\cellcolor{blue!10}{\cmark} & \cellcolor{blue!10}{IN-1K} & \cellcolor{blue!10}{16$\times$288$^2$} & \cellcolor{blue!10}{26} & \cellcolor{blue!10}{112$\times$3$\times$2} & \cellcolor{blue!10}{68.1} & \cellcolor{blue!10}{91.2} \\

             ~ & \cellcolor{blue!10}{\ModelName-M} & \cellcolor{blue!10}{\cmark} & \cellcolor{blue!10}{IN-1K} & \cellcolor{blue!10}{8$\times$224$^2$} & \cellcolor{blue!10}{74} & \cellcolor{blue!10}{101$\times$3$\times$4} & \cellcolor{blue!10}{67.3} & \cellcolor{blue!10}{91.0} \\
             ~ & \cellcolor{blue!10}{\ModelName-M} & \cellcolor{blue!10}{\cmark} & \cellcolor{blue!10}{IN-1K} & \cellcolor{blue!10}{16$\times$224$^2$} & \cellcolor{blue!10}{74} & \cellcolor{blue!10}{202$\times$3$\times$4} & \cellcolor{blue!10}{68.3} & \cellcolor{blue!10}{91.4} \\
             ~ & \cellcolor{blue!10}{\ModelName-M} & \cellcolor{blue!10}{\cmark} & \cellcolor{blue!10}{IN-1K} & \cellcolor{blue!10}{16$\times$288$^2$} & \cellcolor{blue!10}{74} & \cellcolor{blue!10}{333$\times$3$\times$4} & \cellcolor{blue!10}{\textbf{68.4}} & \cellcolor{blue!10}{\textbf{91.6}} \\
             
             \Xhline{1.0pt}
             \multicolumn{9}{l}{\textit{\textbf{\blue{Self-supervised:}} \gray{For UMT, the CLIP-400M is used in pretrained teacher.}}} \\
             \multirow{4}{*}{\textit{\textbf{Trans.}}} & BEVT-B$_{800e}$~\cite{bevt} & \xmark & \scriptsize{IN-1K+K400} & 32$\times$224$^2$ & 88 & 321$\times$3$\times$1 & 70.6 & - \\
             \cline{2-9}
             ~ & VideoMAE-S$_{2400e}$~\cite{videomae} & \cmark &  & 16$\times$224$^2$ & 22 & 57$\times$3$\times$2 & 66.8 & 90.3 \\
             ~ & VideoMAE-B$_{2400e}$~\cite{videomae} & \cmark &  & 16$\times$224$^2$ & 87 & 180$\times$3$\times$2 & \underline{70.8} & 92.4 \\
             ~ & UMT-B$_{800e}$~\cite{umt} & \cmark & \gray{CLIP-400M} & 8$\times$224$^2$ & 87 & 180$\times$3$\times$2 & \underline{70.8} & \underline{92.6} \\

             \hline
             \multirow{3}{*}{\textit{\textbf{SSM}}} & \cellcolor{blue!10}{\ModelName-M$_{800e}$} & \cellcolor{blue!10}{\cmark} & \cellcolor{blue!10}{\gray{CLIP-400M}} & \cellcolor{blue!10}{8$\times$224$^2$} & \cellcolor{blue!10}{74} & \cellcolor{blue!10}{101$\times$3$\times$2} & \cellcolor{blue!10}{70.2} & \cellcolor{blue!10}{92.6} \\
             ~ & \cellcolor{blue!10}{\ModelName-M$_{800e}$} & \cellcolor{blue!10}{\cmark} & \cellcolor{blue!10}{\gray{CLIP-400M}} & \cellcolor{blue!10}{16$\times$224$^2$} & \cellcolor{blue!10}{74} & \cellcolor{blue!10}{202$\times$3$\times$2} & \cellcolor{blue!10}{71.0} & \cellcolor{blue!10}{92.7} \\
             ~ & \cellcolor{blue!10}{\ModelName-M$_{800e}$} & \cellcolor{blue!10}{\cmark} & \cellcolor{blue!10}{\gray{CLIP-400M}} & \cellcolor{blue!10}{16$\times$288$^2$} & \cellcolor{blue!10}{74} & \cellcolor{blue!10}{333$\times$3$\times$2} & \cellcolor{blue!10}{\textbf{71.4}} & \cellcolor{blue!10}{\textbf{92.9}} \\
        \Xhline{1.0pt}	
        \end{tabular}
    }
    \caption{\textbf{Comparison with the state-of-the-art on temporal-related SthSth V2.} 
    ``\textit{iso.}'' means isotropic architecture without downsampling layers.
    Masked modeling~\cite{umt} also works for Mamba, 
    and it performs better than VideoMAE.
    }
    \label{results_ssv2}
    \vspace{-0.6cm}
\end{table}

\subsection{Short-term Video Understanding}
\label{sec:short_term}

\noindent\textbf{Datasets and Settings.} 
We evaluate our \ModelName\  on the popular scene-related Kinetics-400~\cite{k400} and temporal-related Something-Something V2~\cite{sth},
the average video lengths of which are 10s and 4s.
For supervised pretraining,
we fine-tune those models pretrained on ImageNet-1K with the same training strategy as VideoMAE~\cite{videomae}.
Specifically,
for \ModelName-M,
the warmup epoch, total epoch, stochastic depth rate, weight decay are set to 5, 50, 0.8, 0.05 for K400,
and 5, 30, 0.8, 0.05 for SthSth.
For the smaller models,
all the hyper-parameters are the same unless we decrease the stochastic depth rate and increase the training epochs.
Moreover, 
we linearly scale the base learning rates according to the batch size,
which are $2e^{-4}\cdot\frac{batchsize}{256}$ for K400 and $4e^{-4}\cdot\frac{batchsize}{256}$ for SthSth.
As for self-supervised pretraining,
we adopt the training recipe as in UMT~\cite{umt},
employing CLIP-ViT-B~\cite{clip} to distill \ModelName-M over 800 epochs. 
During fine-tuning,
we use similar hyperparameters as mentioned but opt for a small stochastic depth rate and learning rate for both datasets.

\vspace{0.05cm}
\noindent\textbf{Results.}
Table \ref{results_k400} and \ref{results_ssv2} list the results on short-term video datasets.
\textbf{(a) \textit{Supervised}:}
Compared with the purely attention-based methods~\cite{timesformer,vivit},
our SSM-based \ModelName-M secures a notable advantage, 
outperforming ViViT-L~\cite{vivit} by \textbf{+2.0\%} and \textbf{+3.0\%} on the scene-related K400 and the temporally-related SthSthV2 datasets, respectively. 
This improvement comes with significantly reduced computational demands and less pretraining data.
Furthermore, 
\ModelName-M delivers results that are on par with the SOTA UniFormer~\cite{uniformer}, 
which skillfully integrates convolution with attention in a non-isotropic structure.
\textbf{(b) \textit{Self-supervised}:}
The performance of \ModelName\  under masked pretraining surpasses that of the VideoMAE~\cite{videomae}, 
known for its proficiency in fine-grained action. This achievement underscores the potential of our purely SSM-based model in efficiently and effectively understanding short-term videos, highlighting its suitability for both supervised and self-supervised learning paradigms.

\begin{figure}[t]
\vspace{-0.3cm}
\centering
\begin{minipage}{0.3\textwidth}
\centering
    \vspace{0pt}
    \centering
    \setlength\tabcolsep{1pt}
    \resizebox{\textwidth}{!}{
        \begin{tabular}{l|c}
            \textbf{Type} & \textbf{SSV2} \\
            \Xhline{1.0pt}
            \rowcolor{gray!20}
            SF-Bidirectional & \textbf{65.1} \\
            TF-Bidirectional & 62.4 \\
            ST-Bidirectional v1 & 63.9 \\
            ST-Bidirectional v2 & 64.2 \\
            Half-SF + Half-TF & 64.0 \\
            Half-TF + Half-SF & 64.1 \\
            Alternative SF\&TF & \textbf{65.1} \\
        \end{tabular}
    }
    \subcaption{
        \textbf{Scan Type.}
        Spatial-First scan is simple yet effective.
    }
    \label{tab:ablation_scan}
\end{minipage}
\begin{minipage}{0.69\textwidth}

\begin{minipage}{0.49\textwidth}
    \vspace{0pt}
    \centering
    \includegraphics[width=\linewidth]{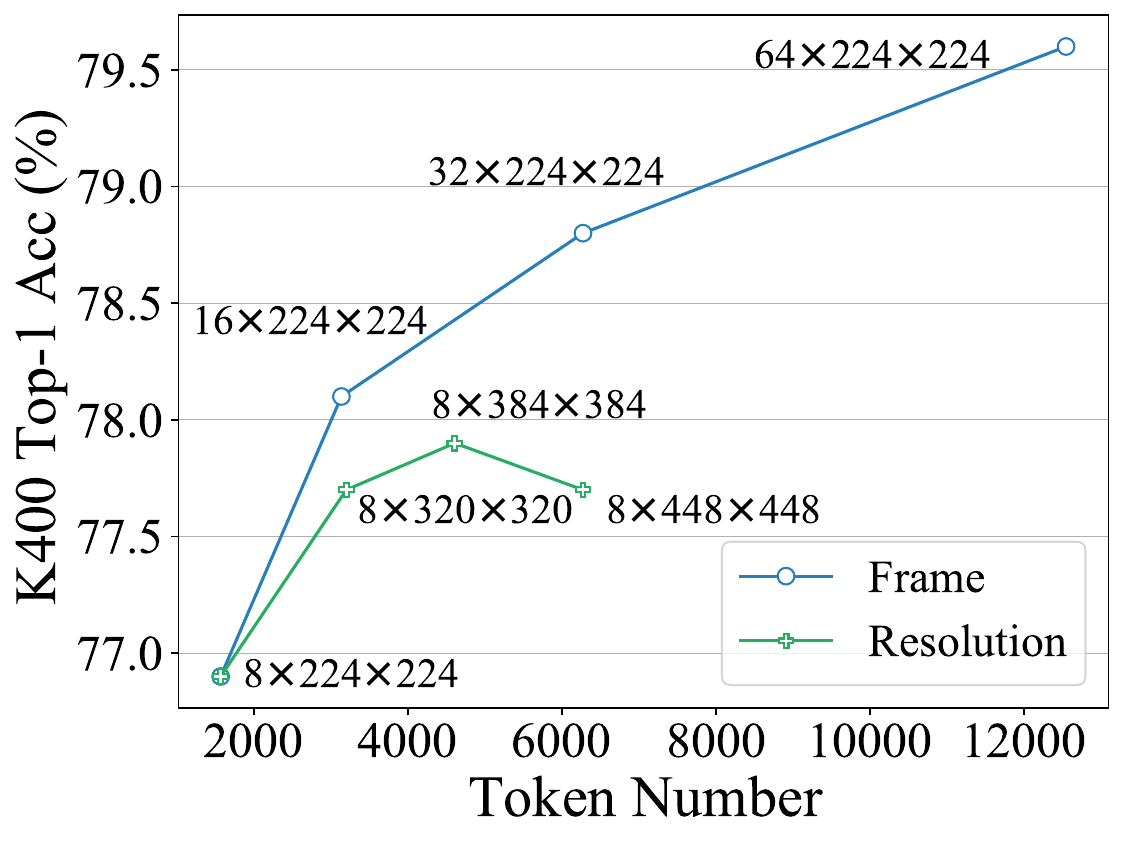}
\end{minipage}
\begin{minipage}{0.49\textwidth}
    \vspace{0pt}
    \centering
    \includegraphics[width=\linewidth]{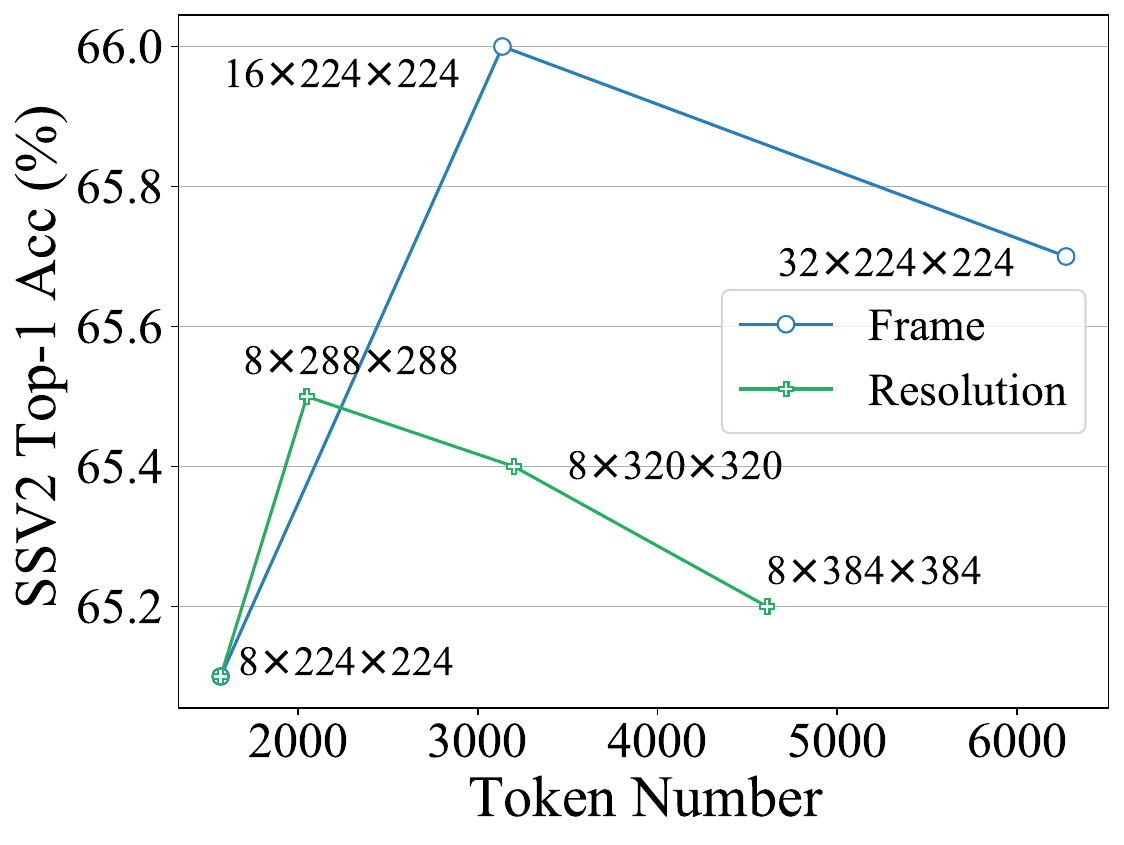} 
\end{minipage}
\subcaption{\textbf{Frame \& Resolution for K400 and SSV2.}}
\end{minipage}
\vspace{-0.3cm}
\caption{\textbf{Ablation studies of scan type, frame and resolution.}
All the models are fine-tuned from \ModelName-Ti pretrained on ImageNet.
} 
\label{tab:ablation_scan_frame_resolution}
\vspace{-0.3cm}
\end{figure}

\vspace{0.05cm}
\noindent\textbf{Ablation Studies.} 
Through comprehensive ablation studies detailed in Fig. \ref{tab:ablation_scan_frame_resolution} and Table \ref{ablation_mask}, 
we explore various aspects of our model.
\textbf{(a) \textit{Scan Type}:}
Among all the methods,
the spatial-first approach emerges as the most effective, 
in contrast, 
the temporal-first strategy is the worst. 
The superiority of the spatial-first method is attributed to its ability to seamlessly leverage 2D pretrained knowledge by scanning frame by frame.
\textbf{(b) \textit{Frame and Resolution}:}
Contrary to findings from ImageNet (see Table \ref{results_imagenet}), 
higher resolution does not uniformly lead to better performance. 
Increasing the number of frames consistently enhances results on the K400 dataset. 
However, 
this is not the case with SthSthV2, 
possibly due to the brief duration of its videos, which may not accommodate longer inputs effectively.
\textbf{(c) \textit{Masked Pretraining}:}
Our findings reveal that row masking, 
being particularly compatible with 1D convolution, outperforms commonly used random and tube masking. 
Clip-row masking excels owing to its higher degree of randomness. 
Moreover, 
attention masking stands out as the most efficient by favoring the preservation of adjacent meaningful content. 
Aligning solely the model's final output proves most effective, 
likely due to architectural differences. 
Lastly, an optimal masking ratio (80\%) combined with stronger regularization significantly benefits \ModelName\  during masked pretraining.

\begin{table}[t]
    \begin{minipage}[t]{0.27\linewidth}
        \vspace{0pt}
        \centering
        \setlength\tabcolsep{8.0pt}
        \resizebox{\textwidth}{!}{
            \begin{tabular}{l|c}
                \textbf{Type} & \textbf{SSV2} \\
                \Xhline{1.0pt}
                Random & 67.4 \\
                Tube & 66.3 \\
                Clip-Row & 68.2 \\
                Frame-Row & 67.8 \\
                \rowcolor{gray!20}
                Attention & \textbf{68.5} \\
            \end{tabular}
        }
        \subcaption{
            \textbf{Mask Type.}
        }
        \label{ablation_mask_type}
    \end{minipage}
    \hspace{0.3mm}
    \ 
    \begin{minipage}[t]{0.25\linewidth}
        \vspace{0pt}
        \centering
        \setlength\tabcolsep{6.5pt}
        \resizebox{\textwidth}{!}{
            \begin{tabular}{l|c}
                \textbf{Layer} & \textbf{SSV2} \\
                \Xhline{1.0pt}
                \rowcolor{gray!20}
                Last 1 & \textbf{68.5} \\
                Last 2 & 68.4 \\
                Last 6 & 68.2 \\
                Last 6$\times$2 & 67.7 \\
            \end{tabular}
        }
        \subcaption{
            \textbf{Alignment Layer.}
        }
        \label{ablation_mask_layer}
    \end{minipage}
    \hspace{0.3mm}
    \ 
    \begin{minipage}[t]{0.22\linewidth}
        \vspace{0pt}
        \centering
        \setlength\tabcolsep{6.5pt}
        \resizebox{\textwidth}{!}{
            \begin{tabular}{l|c}
                \textbf{Ratio} & \textbf{SSV2} \\
                \Xhline{1.0pt}
                50\% & 68.1 \\
                65\% & 68.4 \\
                \rowcolor{gray!20}
                80\% & \textbf{68.5} \\
                90\% & 68.2 \\
            \end{tabular}
        }
        \subcaption{
            \textbf{Mask Ratio.}
        }
        \label{ablation_mask_ratio}
    \end{minipage}
    \hspace{0.3mm}
    \ 
    \begin{minipage}[t]{0.18\linewidth}
        \vspace{0pt}
        \centering
        \setlength\tabcolsep{5.2pt}
        \resizebox{\textwidth}{!}{
            \begin{tabular}{l|c}
                \textbf{DP} & \textbf{SSV2} \\
                \Xhline{1.0pt}
                0.1 & 68.0 \\
                0.2 & 68.2 \\
                0.3 & 68.4 \\
                \rowcolor{gray!20}
                0.4 & \textbf{68.5} \\
            \end{tabular}
        }
        \subcaption{
            \textbf{Droppath.}
        }
        \label{ablation_mask_droppath}
    \end{minipage}
    \vspace{-0.3cm}
    \caption{
        \textbf{Ablation studies of masked pretraining.}
        We adopt CLIP-ViT-B~\cite{clip} as a teacher to distill \ModelName-M for 200 epochs.
    }
    \label{ablation_mask}
    \vspace{-0.5cm}
\end{table}

\subsection{Long-term Video Understanding}

\begin{table}[t]
    \vspace{-0.3cm}
    \centering
    \setlength\tabcolsep{2pt}
    \resizebox{0.95\linewidth}{!}{
        \begin{tabular}{l|c|l|l|l|cc}
        \Xhline{1.0pt}
            \multirow{2}*{\textbf{Method}} & \multirow{2}*{\textit{\textbf{e2e}}} & \multirow{2}*{\textbf{Backbone}} & \multirow{2}*{\textbf{Neck Type}} & \textbf{Pretraining}  & \textbf{BF} & \textbf{COIN} \\
            ~ & ~ & ~ & ~ & \textbf{Dataset} & \textbf{Top-1} & \textbf{Top-1} \\
            \Xhline{0.8pt}
             Timeception~\cite{timeception} & \xmark & 3D-ResNet & Conv. & IN-1K+K400 & 71.3 & - \\
             VideoGraph~\cite{videograph} & \xmark & I3D & Conv.+Atten. & IN-1K+K400 & 69.5 & - \\
             GHRM~\cite{ghrm} & \xmark & I3D & Graph Conv.. & IN-1K+K400 & 75.5 & - \\
             Distant Supervision~\cite{distant} & \xmark & TimeSformer & Atten. w/ KB & IN-21K+HTM & \textbf{89.9} & \textbf{90.0} \\
             ViS4mer~\cite{vis4mer} & \xmark & Swin-B & SSM & IN-21K+K600 & \underline{88.2} & \underline{88.4} \\
             \Xhline{0.8pt}
             Turbo$_{f32}$~\cite{turbo} & \cmark & VideoMAE-B &  & K400 & 86.8 & 82.3 \\
             Turbo$_{f32}$~\cite{turbo} & \cmark & VideoMAE-B &  & K400+HTM-AA & \underline{91.3} & \underline{87.5} \\
             \rowcolor{blue!10}
             \ModelName$_{f32}$ & \cmark & \ModelName-Ti &  & K400 & 94.3 & 86.2 \\
             \rowcolor{blue!10}
             \ModelName$_{f64}$ & \cmark & \ModelName-Ti &  & K400 & 94.3 & 87.0 \\
             \rowcolor{blue!10}
             \ModelName$_{f32}$ & \cmark & \ModelName-S &  & K400 & 95.3 & 88.4 \\
             \rowcolor{blue!10}
             \ModelName$_{f64}$ & \cmark & \ModelName-S &  & K400 & 97.4 & 88.7 \\
             \rowcolor{blue!10}
             \ModelName$_{f32}$ & \cmark & \ModelName-M &  & K400 & 94.8 & 88.3 \\
             \rowcolor{blue!10}
             \ModelName$_{f64}$ & \cmark & \ModelName-M &  & K400 & 95.8 & 89.5 \\
             \rowcolor{blue!10}
             \ModelName$_{f32}$ & \cmark & \ModelName-M\myred{$\dag$} &  & K400 & \textbf{97.9} & 89.6 \\
             \rowcolor{blue!10}
             \ModelName$_{f64}$ & \cmark & \ModelName-M\myred{$\dag$} &  & K400 & 96.9 & \textbf{90.4} \\
        \Xhline{1.0pt}	
        \end{tabular}
    }
    \caption{\textbf{Comparison with the state-of-the-art on Breakfast and COIN.} 
    ``\textit{e2e}'' means end-to-end methods without exhausting feature extraction.
    ``\myred{$\dag$}'' marks the backbone with masked pretraining.
    }
    \label{results_breakfast_coin}
    \vspace{-0.6cm}
\end{table}  

\begin{table}[t]
    \centering
    \setlength\tabcolsep{2pt}
    \resizebox{0.98\linewidth}{!}{
        \begin{tabular}{l|c|l|ccc|cccc|cc}
        \Xhline{1.0pt}
            \multirow{2}*{\textbf{Method}} & \multirow{2}*{\textit{\textbf{e2e}}} & \multirow{2}*{\textit{\textbf{Backbone}}} & \multicolumn{3}{c|}{\textbf{Content($\uparrow$)}} & \multicolumn{4}{c|}{\textbf{Metadata($\uparrow$)}} & \multicolumn{2}{c}{\textbf{User($\downarrow$)}} \\
            ~ & ~ & ~ & \textbf{Rel.} & \textbf{Speak} & \textbf{Scene} & \textbf{Dir.} & \textbf{Genre} & \textbf{Wtr.} & \textbf{Year} & \textbf{Like} & \textbf{View} \\
            \Xhline{0.8pt}
            VideoBERT~\cite{sun2019videobert} & \xmark & S3D & 52.80 & 37.90 & 54.90 & 47.30 & 51.90 & 38.50 & 36.10 & 0.32 & 4.46 \\
            Object Trans.\cite{lvu} & \xmark & ResNet &  53.10 & 39.40 & 56.90 & 51.20 & 54.60 & 34.50 & 39.10 & \textbf{0.23} & \underline{3.55} \\
            LST~\cite{vis4mer} & \xmark & ViT-L &  52.38 & 37.31 &	62.79 &	56.07 &	52.70 &	42.26 &	39.16 &	0.31 &	3.83 \\
            Performer~\cite{vis4mer} & \xmark & ViT-L & 50.00 &	38.80 &	60.46 &	58.87 &	49.45 &	48.21 &	41.25 &	0.31 &	3.93 \\
            Orthoformer~\cite{vis4mer} & \xmark & ViT-L &  50.00 &	39.30 &	66.27 &	55.14 &	\underline{55.79} &	47.02&	43.35 &	0.29 &	3.86 \\
            ViS4mer~\cite{vis4mer} & \xmark & ViT-L &  \underline{57.14} & \textbf{40.79} & \underline{67.44} & \underline{62.61} & 54.71 & \underline{48.80} & \underline{44.75} & \underline{0.26} & 3.63 \\
            \Xhline{0.8pt}
            \rowcolor{blue!10}
            \ModelName$_{f32}$ & \cmark & VM-Ti & \textbf{62.50} & \underline{40.43} & \textbf{70.37} & \textbf{67.29} & \textbf{65.24} & \textbf{52.98} & \textbf{48.23} & \underline{0.26} & \textbf{2.90} \\
        \Xhline{1.0pt}	
        \end{tabular}
    }
    \caption{\textbf{Comparison with the state-of-the-art on LVU.} 
    ``\textit{e2e}'' means end-to-end methods without exhausting feature extraction.
    ``Rel.'', ``Dir.'' and ``Wtr.'' refers to ``Relation'', ``Director'' and ``Writer'', respectively.
    }
    \label{results_lvu}
    \vspace{-0.5cm}
\end{table}

\noindent\textbf{Datasets and Settings.} 
We rigorously assess \ModelName's proficiency in processing long-term videos by leveraging three comprehensive datasets,
\textit{i.e.,}
Breakfast~\cite{breakfast}, COIN~\cite{coin} and Long-form Video Understanding (LVU~\cite{lvu}) benchmark. 
Specifically,
Breakfast comprises 1,712 videos, 
encapsulating 10 intricate cooking activities over 77 hours. 
COIN features 11,827 videos across 180 unique procedural tasks, 
with an average duration of 2.36 minutes. 
The LVU benchmark includes approximately 30K movie clips, 
lasting between 1 to 3 minutes, 
and encompasses nine tasks across 3 primary categories: 
content understanding, metadata prediction, and user engagement. 
For the regression task among these, we evaluate using mean-squared error, 
while for the classification tasks, accuracy is the metric of choice.
In contrast to prior studies~\cite{distant,vis4mer} that rely on features derived from pretrained video models, 
such as Swin-B~\cite{swin} trained on Kinetics-600, 
our method employs end-to-end training as detailed in Section \ref{sec:short_term}. 
Additionally, 
for fair comparisons, 
we fine-tune our models pretrained on K400.

\vspace{0.05cm}
\noindent\textbf{Results.} 
As illustrated in Figure \ref{fig:comparison_memory_speed}, 
the linear complexity of \ModelName\  makes it well-suited for end-to-end training with long-duration videos.
The comparisons in Tables \ref{results_breakfast_coin} and \ref{results_lvu} highlight \ModelName's simplicity and effectiveness against traditional feature-based methods~\cite{vis4mer,distant} on these tasks. 
It yields significant performance improvements, 
achieving SOTA results even with smaller model sizes.
For example, 
\ModelName-Ti shows a notable increase of \textbf{+6.1\%} over ViS4mer using Swin-B features and a \textbf{+3.0\%} uplift against Turbo's multi-modality alignment approach~\cite{turbo}. 
Notably, 
the results underscore the positive impact of the scaling model and frame numbers for long-term tasks. 
In the diverse and challenging set of nine tasks presented by LVU, 
our \ModelName-Ti, fine-tuned in an end-to-end manner, 
delivers outstanding or comparable results to current SOTA methods. 
These outcomes not only highlight VideoMamba's effectiveness but also its great potential for future long-video comprehension.

\subsection{Multi-modality Video Understanding}

\noindent\textbf{Datasets and Settings.} 
Following UMT~\cite{umt},
we utilize WebVid-2M~\cite{bain2021frozen} video-text pairs and CC3M~\cite{cc3m} image-text pairs for joint pretraining with four objectives:
vision-text contrastive learning~\cite{bain2021frozen},
vision-text matching~\cite{li2021align},
masked language modeling~\cite{devlin2018bert} and unmasked token alignment~\cite{umt}.
Initially, 
we mask 50\% image tokens and 80\% video tokens, 
conducting pretraining across 8 frames for 10 epochs. 
Given Mamba's sensitivity to positional information, 
an additional unmasked tuning phase is carried out for one epoch to refine its comprehension further.
For evaluation,
we undertake zero-shot video-text retrieval tasks across five prominent benchmarks, 
including MSRVTT~\cite{msrvtt}, DiDeMo~\cite{didemo}, ActivityNet~\cite{activitynet}, LSMDC~\cite{lsmdc}, and MSVD~\cite{msvd}.
 
\begin{table}[t]
    \vspace{-0.3cm}
    \centering
    \setlength{\tabcolsep}{1.2pt}
    \resizebox{1.0\linewidth}{!}{
        \begin{tabular}{l|l|r|ccc|ccc|ccc|ccc|ccc}
        \Xhline{1.0pt}
        \multirow{2}{*}{\bf Method} & \multirow{2}{*}{\bf BB} & \multirow{2}{*}{\bf \#P} & \multicolumn{3}{c|}{\bf MSRVTT} & \multicolumn{3}{c|}{\bf DiDeMo} & \multicolumn{3}{c|}{\bf ANet} & \multicolumn{3}{c|}{\bf LSMDC} & \multicolumn{3}{c}{\bf MSVD}\\
        & ~ & ~ & @1 & @5 & @10 & @1 & @5 & @10 & @1 & @5 & @10 & @1 & @5 & @10 & @1 & @5 & @10\\
        \Xhline{0.8pt}
        Singularity~\cite{lei2022revealing} & Swin & 5M & 28.4 & 50.2 & 59.5 & \textbf{36.9} & \underline{61.1} & \textbf{69.3} & \underline{30.8} & \underline{55.9} & \underline{66.3}  & - & -  & - & - & -  & - \\
        Frozen~\cite{bain2021frozen} & ViT  & 5M & 18.7 & 39.5 & 51.6 & 20.2 & 46.4 & 58.5 & - & - & - & - & - & - & - & - & - \\
        ALPRO~\cite{li2022align} & ViT  & 5M & 24.1 & 44.7 & 55.4 & 23.8 & 47.3 & 57.9 & - & - & - & - & - & - & - & - & - \\
        BridgeFormer~\cite{bridgeformer} & ViT  & 5M & 26.0 & 46.4 & 56.4 & 25.6 & 50.6 & 61.1 & - & - & - & 12.2 & 25.9 & 32.2 & \textbf{43.6} & \textbf{74.9} & \textbf{84.9} \\
        UMT~\cite{umt} & ViT & 5M & \underline{29.6} & \underline{52.8} & \underline{61.9} & 33.4 & 58.3 & 67.0 & 28.3 & 53.0 & 64.2 & \underline{16.8} & \underline{30.5} & \underline{37.6} & 36.2 & 65.7 & 76.1 \\
        \hline
        \rowcolor{blue!10}
        \ModelName & VM & 5M & \textbf{32.0} & \textbf{53.0} & \textbf{63.8} & \underline{36.6} & \textbf{61.7} & \textbf{70.3} & \textbf{35.9} & \textbf{61.1} & \textbf{72.3} & \textbf{18.0} & \textbf{36.1} & \textbf{43.4} & \underline{38.0} & \underline{68.6} & \underline{79.0} \\
        \Xhline{0.8pt}
        VideoCLIP~\cite{xu2021videoclip}  & S3D & 136M & 10.4 & 22.2 & 30.0 & 16.6 & 46.9 & - & - & - & - & - & - & - & - & - & -\\
        \hline
        VIOLET~\cite{fu2021violet} & Swin & 138M & 25.9 & 49.5 & 59.7 & 23.5 & 49.8 & 59.8 & - & - & - & - & - & - & - & - & - \\
        Singularity~\cite{lei2022revealing} & Swin & 17M & 34.0 & 56.7 & 66.7 & 37.1 & 61.7 & 69.9 & 30.6 & 55.6 & 66.9  & - & -  & - & - & -  & - \\
        OmniVL~\cite{lei2022revealing} & ViT & 17M & 34.6 & 58.4 & 66.6 & 33.3 & 58.7 & 68.5 & - & - & -  & - & -  & - & - & -  & - \\
        UMT~\cite{umt} & ViT & 17M & \underline{35.5} & \textbf{59.3} & \underline{68.6} & 41.9 & 66.7 & 75.0 & 33.8 & 59.1 & 70.4 & 18.1 & 33.1 & 42.2 & 41.4 & 70.6 & 80.1 \\
        UMT~\cite{umt} & ViT & 25M & 35.2 & 57.8 & 66.0 & 41.2 & 65.4 & 74.9 & 35.5 & 60.6 & 71.8 & \underline{19.1} & 33.4 & 42.2 & \underline{42.3} & \textbf{71.7} & \underline{80.8} \\
        \gray{CLIP4Clip~\cite{luo2022clip4clip}} & \gray{ViT} & \gray{400M} & \gray{30.6} & \gray{54.4} & \gray{64.3} & \gray{-} & \gray{-} & \gray{-} & \gray{-} & \gray{-} & \gray{-} & \gray{13.6} & \gray{27.9} & \gray{35.5} & \gray{36.2} & \gray{63.8} & \gray{73.5} \\
        \gray{InternVideo~\cite{Wang2022InternVideoGV}} & \gray{ViT} & \gray{640M} & \gray{40.0} & \gray{65.3} & \gray{74.1} & \gray{31.5} & \gray{57.6} & \gray{68.2} & \gray{30.7} & \gray{57.4} & \gray{70.2} & \gray{17.6} & \gray{32.4} & \gray{40.2} & \gray{43.4} & \gray{69.9} & \gray{79.1} \\
        \hline
        \rowcolor{blue!10}
        \ModelName & VM & 17M & 34.7 & \underline{58.9} & 68.0 & \underline{42.0} & \underline{67.3} & \underline{76.8} & \underline{40.1} & \underline{65.7} & \underline{76.1} & 18.4 & \underline{35.3} & \underline{43.0} & 40.3 & 70.0 & 79.7 \\
        \rowcolor{blue!10}
        \ModelName & VM & 25M & \textbf{35.6} & 58.1 & \textbf{69.5} & \textbf{43.1} & \textbf{68.1} & \textbf{77.7} & \textbf{41.0} & \textbf{67.5} & \textbf{77.8} & \textbf{20.4} & \textbf{37.1} & \textbf{45.7} & \textbf{42.6} & \underline{71.6} & \textbf{81.2} \\
        \Xhline{1.0pt}
        \end{tabular}
    }
    \caption{\textbf{Zero-shot text-to-video retrieval on MSRVTT, DiDeMo, AcitivityNet, LSMDC, and MSVD.}
    ``BB'' means the visual backbone. 
    ``\#P'' refers to the number of pretraining pairs. 
    Models pretrained with large-scale pairs are noted in \gray{gray}.
    }
    \label{tab:results_video_text}
    \vspace{-0.5cm}
\end{table}

\vspace{0.05cm}
\noindent\textbf{Results.} 
As indicated in Table \ref{tab:results_video_text}, 
under the same pretraining corpus and similar training strategies, 
our \ModelName\  achieves superior zero-shot video retrieval performances to UMT~\cite{umt} based on ViT~\cite{vit}.
It underscores Mamba's comparable efficiency and scalability to the ViT in handling multi-modal video tasks.
Notably, 
for datasets featuring longer video lengths (\textit{e.g.}, ANet and DiDeMo) and more complex scenarios (\textit{e.g.}, LSMDC), 
\ModelName\  demonstrates a significant improvement. 
This demonstrates Mamba's aptitude for the demands of cross-modality alignment even in challenging multimodal contexts.

\section{Conclusion}
\label{sec:conclusion}
In this paper,
we propose \ModelName,
a purely SSM-based model for efficient video understanding.
Our extensive experiments demonstrate its 
scalability in the visual domain, 
sensitivity for short-term action recognition,
superiority in long-term video understanding and compatibility with other modalities. 
We hope it can pave the way for future model design for long-video comprehension.

\vspace{0.05cm}
\noindent\textbf{Limitations.}
Due to resource constraints, 
we have not yet fully validated the scalability of \ModelName, 
such as extending \ModelName\  to larger sizes (\textit{e.g}., \ModelName-g), 
incorporating additional modalities (\textit{e.g}.,  audio), 
and integrating with large language models for hour-level video understanding. 
Despite these limitations, 
our findings confirm \ModelName's promising potential and we plan to conduct thorough explorations of its capabilities in the future.
\clearpage

%
%
\bibliographystyle{splncs04}
\bibliography{main}

\clearpage
\appendix
\renewcommand{\thetable}{\Roman{table}}
\renewcommand{\thefigure}{\Roman{figure}}


\title{\logo\ModelName: State Space Model for Efficient Video Understanding}
\titlerunning{VideoMamba}
\author{\Large{Appendix}}
\authorrunning{K. Li et al.}
\institute{
\large{\url{https://github.com/OpenGVLab/VideoMamba}}
}

\maketitle

\begin{table}[t]
    \centering
    \setlength\tabcolsep{2pt}
    \resizebox{0.95\linewidth}{!}{
        \begin{tabular}{l|l|c|l|r|r|r|cc}
        \Xhline{1.0pt}
            \multirow{2}*{\textbf{Arch.}} & \multirow{2}*{\textbf{Model}} & \multirow{2}*{\textit{\textbf{iso.}}} & \textbf{Extra} & \textbf{Input} & \textbf{\#Param} & \textbf{FLOPs} & \multicolumn{2}{c}{\textbf{K400}} \\
            ~ & ~ & ~ & \textbf{Data} & \textbf{Size} & \textbf{(M)} & \textbf{(G)} & \textbf{Top-1} & \textbf{Top-5} \\
            \Xhline{0.8pt}
             \multicolumn{9}{l}{\textit{\textbf{\blue{Supervised:}} \gray{Those models with extra data are under supervised training.}}} \\
             \Xhline{0.8pt}
             \multirow{4}{*}{\textit{\textbf{Trans.}}} & STAM~\cite{stam} & \cmark & IN-21K & 64$\times$224$^2$ & 121 & 1040$\times$1$\times$1 & 79.2 & - \\
             ~ & TimeSformer-L~\cite{timesformer} & \cmark & IN-21K & 96$\times$224$^2$ & 121 & 2380$\times$3$\times$1 & 80.7 & 94.7 \\
             ~ & ViViT-L~\cite{vivit} & \cmark & IN-21K & 16$\times$224$^2$ & 311 & 3992$\times$3$\times$4 & \underline{81.3} & 94.7 \\
             ~ & Mformer-HR~\cite{motionformer} & \cmark & IN-21K & 16$\times$336$^2$ & 311 & 959$\times$3$\times$10 & 81.1 & \underline{95.2} \\
             \hline
             \multirow{15}{*}{\textit{\textbf{SSM}}} & \cellcolor{blue!10}{\ModelName-Ti} & \cellcolor{blue!10}{\cmark} & \cellcolor{blue!10}{IN-1K} & \cellcolor{blue!10}{8$\times$224$^2$} & \cellcolor{blue!10}{7} & \cellcolor{blue!10}{9$\times$3$\times$4} & \cellcolor{blue!10}{76.9} & \cellcolor{blue!10}{92.9} \\
             ~ & \cellcolor{blue!10}{\ModelName-Ti} & \cellcolor{blue!10}{\cmark} & \cellcolor{blue!10}{IN-1K} & \cellcolor{blue!10}{16$\times$224$^2$} & \cellcolor{blue!10}{7} & \cellcolor{blue!10}{17$\times$3$\times$4} & \cellcolor{blue!10}{78.1} & \cellcolor{blue!10}{93.5} \\
             ~ & \cellcolor{blue!10}{\ModelName-Ti} & \cellcolor{blue!10}{\cmark} & \cellcolor{blue!10}{IN-1K} & \cellcolor{blue!10}{32$\times$224$^2$} & \cellcolor{blue!10}{7} & \cellcolor{blue!10}{34$\times$3$\times$4} & \cellcolor{blue!10}{78.8} & \cellcolor{blue!10}{93.9} \\
             ~ & \cellcolor{blue!10}{\ModelName-Ti} & \cellcolor{blue!10}{\cmark} & \cellcolor{blue!10}{IN-1K} & \cellcolor{blue!10}{64$\times$224$^2$} & \cellcolor{blue!10}{7} & \cellcolor{blue!10}{69$\times$3$\times$4} & \cellcolor{blue!10}{79.6} & \cellcolor{blue!10}{94.2} \\
             ~ & \cellcolor{blue!10}{\ModelName-Ti} & \cellcolor{blue!10}{\cmark} & \cellcolor{blue!10}{IN-1K} & \cellcolor{blue!10}{64$\times$384$^2$} & \cellcolor{blue!10}{7} & \cellcolor{blue!10}{202$\times$3$\times$4} & \cellcolor{blue!10}{80.3} & \cellcolor{blue!10}{94.8} \\

             ~ & \cellcolor{blue!10}{\ModelName-S} & \cellcolor{blue!10}{\cmark} & \cellcolor{blue!10}{IN-1K} & \cellcolor{blue!10}{8$\times$224$^2$} & \cellcolor{blue!10}{26} & \cellcolor{blue!10}{34$\times$3$\times$4} & \cellcolor{blue!10}{79.3} & \cellcolor{blue!10}{94.2} \\
             ~ & \cellcolor{blue!10}{\ModelName-S} & \cellcolor{blue!10}{\cmark} & \cellcolor{blue!10}{IN-1K} & \cellcolor{blue!10}{16$\times$224$^2$} & \cellcolor{blue!10}{26} & \cellcolor{blue!10}{68$\times$3$\times$4} & \cellcolor{blue!10}{80.8} & \cellcolor{blue!10}{94.8} \\
             ~ & \cellcolor{blue!10}{\ModelName-S} & \cellcolor{blue!10}{\cmark} & \cellcolor{blue!10}{IN-1K} & \cellcolor{blue!10}{32$\times$224$^2$} & \cellcolor{blue!10}{26} & \cellcolor{blue!10}{135$\times$3$\times$4} & \cellcolor{blue!10}{81.5} & \cellcolor{blue!10}{95.2} \\
             ~ & \cellcolor{blue!10}{\ModelName-S} & \cellcolor{blue!10}{\cmark} & \cellcolor{blue!10}{IN-1K} & \cellcolor{blue!10}{64$\times$224$^2$} & \cellcolor{blue!10}{26} & \cellcolor{blue!10}{271$\times$3$\times$4} & \cellcolor{blue!10}{81.8} & \cellcolor{blue!10}{95.3} \\
             ~ & \cellcolor{blue!10}{\ModelName-S} & \cellcolor{blue!10}{\cmark} & \cellcolor{blue!10}{IN-1K} & \cellcolor{blue!10}{64$\times$384$^2$} & \cellcolor{blue!10}{26} & \cellcolor{blue!10}{395$\times$3$\times$4} & \cellcolor{blue!10}{82.7} & \cellcolor{blue!10}{95.6} \\

             ~ & \cellcolor{blue!10}{\ModelName-M} & \cellcolor{blue!10}{\cmark} & \cellcolor{blue!10}{IN-1K} & \cellcolor{blue!10}{8$\times$224$^2$} & \cellcolor{blue!10}{74} & \cellcolor{blue!10}{101$\times$3$\times$4} & \cellcolor{blue!10}{80.6} & \cellcolor{blue!10}{94.6} \\
             ~ & \cellcolor{blue!10}{\ModelName-M} & \cellcolor{blue!10}{\cmark} & \cellcolor{blue!10}{IN-1K} & \cellcolor{blue!10}{16$\times$224$^2$} & \cellcolor{blue!10}{74} & \cellcolor{blue!10}{202$\times$3$\times$4} & \cellcolor{blue!10}{81.9} & \cellcolor{blue!10}{95.4} \\
             ~ & \cellcolor{blue!10}{\ModelName-M} & \cellcolor{blue!10}{\cmark} & \cellcolor{blue!10}{IN-1K} & \cellcolor{blue!10}{32$\times$224$^2$} & \cellcolor{blue!10}{74} & \cellcolor{blue!10}{403$\times$3$\times$4} & \cellcolor{blue!10}{82.4} & \cellcolor{blue!10}{95.7} \\
             ~ & \cellcolor{blue!10}{\ModelName-M} & \cellcolor{blue!10}{\cmark} & \cellcolor{blue!10}{IN-1K} & \cellcolor{blue!10}{64$\times$224$^2$} & \cellcolor{blue!10}{74} & \cellcolor{blue!10}{806$\times$3$\times$4} & \cellcolor{blue!10}{82.8} & \cellcolor{blue!10}{96.0} \\
             ~ & \cellcolor{blue!10}{\ModelName-M} & \cellcolor{blue!10}{\cmark} & \cellcolor{blue!10}{IN-1K} & \cellcolor{blue!10}{64$\times$384$^2$} & \cellcolor{blue!10}{74} & \cellcolor{blue!10}{2368$\times$3$\times$4} & \cellcolor{blue!10}{\textbf{83.3}} & \cellcolor{blue!10}{\textbf{96.1}} \\
             
             \Xhline{1.0pt}
             \multicolumn{9}{l}{\textit{\textbf{\blue{Self-supervised:}} \gray{For UMT, the CLIP-400M is used in pretrained teacher.}}} \\
             \multirow{4}{*}{\textit{\textbf{Trans.}}} & ST-MAE-B$_{1600e}$~\cite{st_mae} & \cmark &  & 16$\times$224$^2$ & 87 & 180$\times$3$\times$7 & 81.3 & 94.9 \\
             ~ & VideoMAE-S$_{2400e}$~\cite{videomae} & \cmark &  & 16$\times$224$^2$ & 22 & 57$\times$3$\times$5 & 79.0 & 93.8 \\
             ~ & VideoMAE-B$_{1600e}$~\cite{videomae} & \cmark &  & 16$\times$224$^2$ & 87 & 180$\times$3$\times$5 & 81.5 & 95.1 \\
             ~ & UMT-B$_{800e}$~\cite{umt} & \cmark & \gray{CLIP-400M} & 8$\times$224$^2$ & 87 & 180$\times$3$\times$5 & \textbf{85.7} & \textbf{97.0} \\

             \hline
             \multirow{5}{*}{\textit{\textbf{SSM}}} & \cellcolor{blue!10}{\ModelName-M$_{800e}$} & \cellcolor{blue!10}{\cmark} & \cellcolor{blue!10}{\gray{CLIP-400M}} & \cellcolor{blue!10}{8$\times$224$^2$} & \cellcolor{blue!10}{74} & \cellcolor{blue!10}{101$\times$3$\times$4} & \cellcolor{blue!10}{82.0} & \cellcolor{blue!10}{95.4} \\
             ~ & \cellcolor{blue!10}{\ModelName-M$_{800e}$} & \cellcolor{blue!10}{\cmark} & \cellcolor{blue!10}{\gray{CLIP-400M}} & \cellcolor{blue!10}{16$\times$224$^2$} & \cellcolor{blue!10}{74} & \cellcolor{blue!10}{202$\times$3$\times$4} & \cellcolor{blue!10}{83.4} & \cellcolor{blue!10}{95.9} \\
             ~ & \cellcolor{blue!10}{\ModelName-M$_{800e}$} & \cellcolor{blue!10}{\cmark} & \cellcolor{blue!10}{\gray{CLIP-400M}} & \cellcolor{blue!10}{32$\times$224$^2$} & \cellcolor{blue!10}{74} & \cellcolor{blue!10}{403$\times$3$\times$4} &\cellcolor{blue!10}{83.9} & \cellcolor{blue!10}{96.2} \\
             ~ & \cellcolor{blue!10}{\ModelName-M$_{800e}$} & \cellcolor{blue!10}{\cmark} & \cellcolor{blue!10}{\gray{CLIP-400M}} & \cellcolor{blue!10}{64$\times$224$^2$} & \cellcolor{blue!10}{74} & \cellcolor{blue!10}{806$\times$3$\times$4} &\cellcolor{blue!10}{84.3} & \cellcolor{blue!10}{96.6} \\
             ~ & \cellcolor{blue!10}{\ModelName-M$_{800e}$} & \cellcolor{blue!10}{\cmark} & \cellcolor{blue!10}{\gray{CLIP-400M}} & \cellcolor{blue!10}{64$\times$384$^2$} & \cellcolor{blue!10}{74} & \cellcolor{blue!10}{2368$\times$3$\times$4} &\cellcolor{blue!10}{\underline{85.0}} & \cellcolor{blue!10}{\underline{96.9}} \\
        \Xhline{1.0pt}	
        \end{tabular}
    }
    \caption{\textbf{More results on scene-related Kinetics-400.} 
    ``\textit{iso.}'' means isotropic architecture without downsampling layers.
    }
    \label{more_results_k400}
    \vspace{-0.5cm}
\end{table}

\section{More Results}

In Table~\ref{more_results_k400}, 
we present additional results on the Kinetics-400 dataset. 
These results clearly demonstrate that our SSM-based model outperforms all previous attention-based methods. 
We observe consistent performance improvements with increasing resolution and frame count.

\begin{table}[t!]
    \centering
    \setlength\tabcolsep{11pt}
    \resizebox{1.0\linewidth}{!}{
        \begin{tabular}{l|cc|c}
        \multirow{2}*{\textbf{config}} &
        \multicolumn{2}{c|}{\textbf{Single-Modality}} & \multicolumn{1}{c}{\textbf{Multi-Modality}} \\
        ~ & \textbf{SthSthV2} & \textbf{K400} & \textbf{5M \& 17M \& 25M} \\
        \Xhline{1.0pt}
        optimizer & \multicolumn{2}{c|}{\textit{AdamW}} & \textit{AdamW} \\ 
        optimizer momentum & \multicolumn{2}{c|}{$\beta_1, \beta_2{=}0.9, 0.95$} &  $\beta_1, \beta_2{=}0.9, 0.999$ \\
        weight decay & \multicolumn{2}{c|}{0.05} & 0.05 \\
        learning rate schedule & \multicolumn{2}{c|}{\textit{cosine decay}} & \textit{cosine decay} \\
        learning rate & \multicolumn{2}{c|}{1.2e-3} & 4e-4 \\
        minimal learning rate & \multicolumn{2}{c|}{1e-5} & 4e-6 \\
        batch size & \multicolumn{2}{c|}{2048} &  2048 $\mathbf I$, 2048 $\mathbf V$ \\
        warmup epochs & \multicolumn{2}{c|}{40} & 1 \\
        total epochs &  \multicolumn{2}{c|}{800} & 10 \\
        mask ratio & \multicolumn{2}{c|}{80\%} & 50\% $\mathbf I$, 80\% $\mathbf V$, 50\% $\mathbf T$ \\
        input frame & \multicolumn{2}{c|}{8} & 8\\
        drop path & \multicolumn{2}{c|}{0.4} & 0.25 \\
        flip augmentation & \textit{no} & \textit{yes} & \textit{yes} \\
        augmentation & \multicolumn{2}{c|}{\textit{MultiScaleCrop}[0.66, 0.75, 0.875, 1]} & \textit{MultiScaleCrop}[0.5, 1] \\
        \end{tabular}
    }
    \caption{
        \textbf{Masked pre-training settings.}
        $\mathbf I$-image, $\mathbf V$-video, $\mathbf T$-text.
    }
    \label{tab:hyperparameters_pretraining} 
\end{table}

\begin{table}[t]
    \centering
    \setlength\tabcolsep{18pt}
    \resizebox{1.0\linewidth}{!}{
        \begin{tabular}{l|ccc}
        \textbf{config} & \textbf{224$\times$224} & \textbf{448$\times$448} & \textbf{512$\times$512} \\
        \Xhline{1.0pt}
        optimizer & \multicolumn{3}{c}{\textit{AdamW}} \\ 
        optimizer momentum & \multicolumn{3}{c}{$\beta_1, \beta_2{=}0.9, 0.999$}  \\
        weight decay & 0.1\gray{(Ti)}, 0.05\gray{(S,M)} & 1e-8 & 1e-8 \\
        learning rate schedule & \multicolumn{3}{c}{\textit{cosine decay}} \\
        base learning rate & 5e-4 & 5e-6 & 5e-6 \\
        minimal learning rate & 1e-5 & 5e-6 & 5e-6 \\
        base batch size & \multicolumn{3}{c}{512} \\
        repeated augmentation & \multicolumn{3}{c}{\textit{no}\gray{(Ti)}, \textit{yes}\gray{(S,M)}} \\
        warmup epochs & 5\gray{(Ti,S)}, 30\gray{(M)} & 5 & 2 \\
        total epochs &  300 & 30 & 10 \\
        drop path & \multicolumn{3}{c}{0\gray{(Ti)}, 0.15\gray{(S)}, 0.5\gray{(M)}} \\
        label smoothing & \multicolumn{3}{c}{0.1} \\
        cutmix & \multicolumn{3}{c}{1.0} \\
        augmentation & \multicolumn{3}{c}{\textit{RandAug}(7, 0.25)\gray{(Ti)}, \textit{RandAug}(9, 0.5)\gray{(S,M)}} \\
        \end{tabular}
    }
    \caption{
        \textbf{Training settings for ImageNet-1K.}
    }
    \label{tab:hyperparameters_finetuning_in1k} 
\end{table}

\begin{table}[t]
    \centering
    \setlength\tabcolsep{24pt}
    \resizebox{1.0\linewidth}{!}{
        \begin{tabular}{l|cc}
        \textbf{config} & \textbf{224$\times$224} & \textbf{384$\times$384} \\
        \Xhline{1.0pt}
        optimizer & \multicolumn{2}{c}{\textit{AdamW}} \\ 
        optimizer momentum & \multicolumn{2}{c}{$\beta_1, \beta_2{=}0.9, 0.999$}  \\
        weight decay & 0.1\gray{(Ti)}, 0.05\gray{(S,M,M\red{$\dag$})} & 1e-8 \\
        learning rate schedule & \multicolumn{2}{c}{\textit{cosine decay}} \\
        base learning rate & 4e-4\gray{(Ti,S)}, 2e-4\gray{(M)}, 1e-4\gray{(M\red{$\dag$})} & 5e-6 \\
        minimal learning rate & \multicolumn{2}{c}{1e-6} \\
        base batch size & \multicolumn{2}{c}{256} \\
        repeated augmentation & \multicolumn{2}{c}{2} \\
        warmup epochs & 5 & 2 \\
        total epochs &  70\gray{(Ti)}, \gray{50(S,M)}, \gray{45(M\red{$\dag$})} & 10 \\
        drop path & \multicolumn{2}{c}{0.1\gray{(Ti)}, 0.35\gray{(S)}, 0.8\gray{(M)}, 0.4\gray{(M\red{$\dag$})}} \\
        layer-wise lr decay & \multicolumn{2}{c}{0.75\gray{(S,M,M\red{$\dag$})}, 0.8\gray{(M\red{$\dag$})}} \\
        flip augmentation & \multicolumn{2}{c}{\textit{yes}} \\
        label smoothing & \multicolumn{2}{c}{0.1} \\
        cutmix& \multicolumn{2}{c}{1.0} \\
        augmentation & \multicolumn{2}{c}{\textit{RandAug}(7, 0.25)\gray{(Ti)}, \textit{RandAug}(9, 0.5)\gray{(S,M,M\red{$\dag$})}} \\
        \end{tabular}
    }
    \caption{
        \textbf{Training settings for Kinetics-400.}
        ``\myred{$\dag$}'' means masked pretraining.
    }
    \label{tab:hyperparameters_finetuning_k400} 
\end{table}

\begin{table}[t]
    \centering
    \setlength\tabcolsep{22pt}
    \resizebox{1.0\linewidth}{!}{
        \begin{tabular}{l|cc}
        \textbf{config} & \textbf{224$\times$224} & \textbf{288$\times$288} \\
        \Xhline{1.0pt}
        optimizer & \multicolumn{2}{c}{\textit{AdamW}} \\ 
        optimizer momentum & \multicolumn{2}{c}{$\beta_1, \beta_2{=}0.9, 0.999$}  \\
        weight decay & 0.1\gray{(Ti)}, 0.05\gray{(S,M,M\red{$\dag$})} & 1e-8 \\
        learning rate schedule & \multicolumn{2}{c}{\textit{cosine decay}} \\
        base learning rate & 4e-4\gray{(Ti,S,M)} 1e-4\gray{(M\red{$\dag$})} & 5e-6 \\
        minimal learning rate & \multicolumn{2}{c}{1e-6} \\
        base batch size & \multicolumn{2}{c}{256} \\
        repeated augmentation & \multicolumn{2}{c}{2} \\
        warmup epochs & 5 & 2 \\
        total epochs &  35\gray{(Ti)}, \gray{30(S,M,M\red{$\dag$})} & 10 \\
        drop path & \multicolumn{2}{c}{0.1\gray{(Ti)}, 0.35\gray{(S)}, 0.8\gray{(M)}, 0.4\gray{(M\red{$\dag$})}} \\
        layer-wise lr decay & \multicolumn{2}{c}{0.75\gray{(S,M,M\red{$\dag$})}, 0.8\gray{(M\red{$\dag$})}} \\
        flip augmentation & \multicolumn{2}{c}{\textit{no}} \\
        label smoothing & \multicolumn{2}{c}{0.1} \\
        cutmix& \multicolumn{2}{c}{1.0} \\
        augmentation & \multicolumn{2}{c}{\textit{RandAug}(7, 0.25)\gray{(Ti)}, \textit{RandAug}(9, 0.5)\gray{(S,M,M\red{$\dag$})}} \\
        \end{tabular}
    }
    \caption{
        \textbf{Training settings for SthSthV2.}
        ``\myred{$\dag$}'' means masked pretraining.
    }
    \label{tab:hyperparameters_finetuning_ssv2} 
\end{table}

\begin{table}[t]
    \centering
    \setlength\tabcolsep{16pt}
    \resizebox{1.0\linewidth}{!}{
        \begin{tabular}{l|cc}
        \textbf{config} & \textbf{BreakFast \& LVU} & \textbf{COIN} \\
        \Xhline{1.0pt}
        optimizer & \multicolumn{2}{c}{\textit{AdamW}} \\ 
        optimizer momentum & \multicolumn{2}{c}{$\beta_1, \beta_2{=}0.9, 0.999$}  \\
        weight decay & \multicolumn{2}{c}{0.1\gray{(Ti)}, 0.05\gray{(S,M,M\red{$\dag$})}} \\
        learning rate schedule & \multicolumn{2}{c}{\textit{cosine decay}} \\
        base learning rate & \multicolumn{2}{c}{2e-4} \\
        minimal learning rate & \multicolumn{2}{c}{1e-6} \\
        base batch size & \multicolumn{2}{c}{256} \\
        repeated augmentation & \multicolumn{2}{c}{2} \\
        warmup epochs & \multicolumn{2}{c}{5} \\
        total epochs &  70\gray{(Ti)}, \gray{50(S,M)}, \gray{45(M\red{$\dag$})} & 40\gray{(Ti)}, \gray{35(S)}, \gray{30(M,M\red{$\dag$})} \\
        drop path & \multicolumn{2}{c}{0.1\gray{(Ti)}, 0.35\gray{(S)}, 0.8\gray{(M)}, 0.4\gray{(M\red{$\dag$})}} \\
        layer-wise lr decay & \multicolumn{2}{c}{0.75\gray{(S,M,M\red{$\dag$})}, 0.8\gray{(M\red{$\dag$})}} \\
        flip augmentation & \multicolumn{2}{c}{\textit{yes}} \\
        label smoothing & \multicolumn{2}{c}{0.1} \\
        cutmix& \multicolumn{2}{c}{1.0} \\
        augmentation & \multicolumn{2}{c}{\textit{RandAug}(7, 0.25)\gray{(Ti)}, \textit{RandAug}(9, 0.5)\gray{(S,M,M\red{$\dag$})}} \\
        \end{tabular}
    }
    \caption{
        \textbf{Training settings for Breaskfast, COIN and LVU.}
        ``\myred{$\dag$}'' means masked pretraining.
        We directly sample the frames from the raw video sparsely.
    }
    \label{tab:hyperparameters_finetuning_bcl} 
    \vspace{0.3cm}
\end{table}

\section{More Implementation Details}

\subsection{Training Details}

We sparsely sample frames from the raw videos as in TSN~\cite{tsn} for all the datasets. 
Table~\ref{tab:hyperparameters_pretraining} details the masked pretraining hyperparameters.
For the unmasked multi-modality pretraining, we load the pretrained model and train it for an additional epoch with a learning rate of 8e-5. 
Moreover,
Tables~\ref{tab:hyperparameters_finetuning_in1k},~\ref{tab:hyperparameters_finetuning_k400},~\ref{tab:hyperparameters_finetuning_ssv2}, and~\ref{tab:hyperparameters_finetuning_bcl} show the training details for the different datasets used for fine-tuning.

\subsection{Dataset Descriptions}

\begin{table*}[tp]
    \centering
    \setlength\tabcolsep{5.0pt}
    \resizebox{1.0\textwidth}{!}{
        \begin{tabular}{lccc}
        \toprule
        \textbf{Dataset} & \textbf{\#image/video} & \textbf{\#text} & \textbf{Type} \\
        \midrule
        COCO & 113K & 567K & image \\
        Visual Genome & 100K & 768K & image \\
        SBU Captions & 860K & 860K & image \\
        CC3M  & 2.88M & 2.88M & image \\
        CC12M  & 11.00M & 11.00M & image \\
        WebVid-2M  & 2.49M & 2.49M & video \\
        WebVid-10M & 10.73M & 10.73M & video \\
        \midrule
        5M corpus = CC3M$+$WebVid-2M & 5.37M & 5.37M & video+image \\
        17M corpus = 5M$+$COCO$+$VG$+$SBU$+$CC12M & 17.44M & 18.57M & video+image \\
        25M corpus = 17M$+$WebVid-10M$-$WebVid-2M & 25.68M & 26.81M & video+image \\
        \bottomrule
        \end{tabular}
    }
    \caption{\textbf{Statistics of multi-modality datasets.}
    }
    \label{tab:statics_pretrain}
    \vspace{-0.1cm}
\end{table*}

\begin{table*}[tp]
    \centering
    \setlength\tabcolsep{4.0pt}
    \resizebox{1.0\textwidth}{!}{
        \begin{tabular}{lrrrrrrr}
        \toprule
        \multirow{2}{*}{ \textbf{Dataset} } & \multicolumn{3}{c}{\textbf{\#video}} & \multicolumn{3}{c}{\textbf{\#text}} & \textbf{Avg Video} \\  \cmidrule{2-4} \cmidrule(l){5-7}
        & \textbf{Train} & \textbf{Val} & \textbf{Test} & \textbf{Train} & \textbf{Val} & \textbf{Test} & \textbf{Length (s)} \\ 
        \midrule
        \multicolumn{3}{l}{\textit{\textbf{Image Classification}}} & & & & & \\
        ImageNet-1K & 1,281,167 & 50,000 & 100,000 & - & - & - & - \\
        \midrule
        \multicolumn{3}{l}{\textit{\textbf{Short-term Action Recognition}}} & & & & & \\
        Kinetics-400 & 240,436 & 19,787 & - & - & - & - & 10 \\
        Something-Something V2 & 168,913 & 24,777 & - & - & - & - & 4 \\
        \midrule
        \multicolumn{3}{l}{\textit{\textbf{Long-term Action Recognition}}} & & & & & \\
        Breakfast & 1,577 & - & 410 & - & - & - & 137 \\
        COIN & 9,026 & - & 2,796 & - & - & - & 142 \\
        LVU & 7,619 & 1,666 & 1,551 & - & - & - & 134 \\
        \ \ \textit{Relation} & 138 & 49 & 41 & - & - & - & 127 \\
        \ \ \textit{Speak} & 871 & 196 & 188 & - & - & - & 133 \\
        \ \ \textit{Scene} & 514 & 107 & 81 & - & - & - & 132 \\
        \ \ \textit{Director} & 680 & 163 & 107 & - & - & - & 137 \\
        \ \ \textit{Genre} & 2807 & 569 & 584 & - & - & - & 130 \\
        \ \ \textit{Writer} & 748 & 174 & 168 & - & - & - & 142 \\
        \ \ \textit{Year} & 725 & 163 & 141 & - & - & - & 133 \\
        \ \ \textit{Like} & 658 & 142 & 139 & - & - & - & 159 \\
        \ \ \textit{View} & 478 & 103 & 102 & - & - & - & 112 \\
        \midrule
        \multicolumn{3}{l}{\textit{\textbf{Video-Text Retrieval}}} & & & & & \\
        MSRVTT & 7,010 & - & 1,000 & 140,200 & - & 1,000 & 15 \\
        DiDeMo & 8,496 & 1,094 & 1,036 & 8,496 & 1,094 & 1,036 & 29 \\
        ActivityNet & 10,009 & 4,917 & - & 10,009 & 4,917 & - & 180 \\
        LSMDC & 101,055 & - & 1,000 & 101,055 & - & 1,000 & 5 \\
        MSVD & 1,200 & 100 & 670 & 1,200 & 100 & 670 & 15 \\
        \bottomrule
        \end{tabular}
    }
    \caption{\textbf{Statistics of single-modality datasets.}
    }
    \label{tab:statics_downstream}
    \vspace{-0.1cm}
\end{table*}

We show the statistics of multi-modality datasets in Table \ref{tab:statics_pretrain},
and single-modality datasets in Table \ref{tab:statics_downstream}.
\end{document}